\documentclass[letterpaper,10pt,journal]{IEEEtran}
\IEEEoverridecommandlockouts

\usepackage{amsthm}
\usepackage{times}
\usepackage[hidelinks,bookmarks=true]{hyperref}
\usepackage{xcolor}
\usepackage{amsmath,amssymb,amsfonts,bm}
\usepackage{dsfont}
\usepackage{graphicx}
\usepackage{booktabs}
\usepackage[ruled,vlined,linesnumbered,noend]{algorithm2e}
\usepackage{multirow}
\usepackage{mathabx}
\usepackage{array}
\usepackage{tabularx}
\usepackage[T1]{fontenc}
\usepackage{enumitem}

\setlist{nosep,leftmargin=*}

\theoremstyle{definition}

\newtheorem{remark}{Remark}

\title{FlexWorm: Primitive-augmented Hybrid Contact-motion Planning for Suction-based Multi-segment Deformable Robots}

\author{Zili Tang, Tiecheng Guo, Qinyue Zhang, and Meng Guo%
\thanks{The authors are with the School of Advanced Manufacturing and Robotics, 
Peking University, Beijing 100871, China. 
Corresponding author: Meng Guo ({\tt\footnotesize meng.guo@pku.edu.cn}).
This work was supported by the National Key Research and Development Program of China under grant 2023YFB4706500;
the National Natural Science Foundation of China under grants U2241214 and T2121002.
Accepted for publication in IEEE Robotics and Automation Letters (RA-L).}}

\makeatletter
\def\@IEEEpubidpullup{1.5\baselineskip}
\makeatother
\IEEEpubid{%
\parbox{\textwidth}{\centering\scriptsize
\copyright\ 2026 IEEE. Personal use of this material is permitted. Permission from IEEE must be obtained for all other uses, in any current or future media, including reprinting/republishing this material for advertising or promotional purposes, creating new collective works, for resale or redistribution to servers or lists, or reuse of any copyrighted component of this work in other works.}}

\begin{document}
\maketitle

\begin{abstract}
Multi-segment suction-based soft robots are promising for inspection and maintenance in confined or fragile environments,
but existing approaches still depend heavily on manually designed gaits and environment-specific motion scripts.
This work presents a planning framework for serial multi-segment soft robots with deformable body segments and boundary suction pads.
The formulation targets full 3D navigation on complex surfaces and explicitly handles discrete adhesion switching and continuous body deformation
under geometric, collision, and quasi-static feasibility constraints, while remaining agnostic to the specific actuation realization used to produce segment deformation.
Its core, block-wise IK hybrid search (IKHS), performs best-first search over feasible adhesion transitions
while solving inverse kinematics only on induced free blocks.
On top of IKHS, primitive-augmented hybrid search (PaHS) uses a learned observation--primitive embedding
to retrieve short validated motion segments for fast local proposal,
with fallback to standard IKHS branching when retrieval fails.
In simulation, the framework consistently outperforms controlled baselines in planning success,
transition quality, and efficiency across diverse terrains.
PaHS matches IKHS in success rate while substantially reducing planning time.
Repeated hardware experiments on a pneumatic multi-segment soft robot further demonstrate executability
and online recovery under actuation and adhesion uncertainty.
\end{abstract}


\section{Introduction}\label{sec:intro}

\IEEEPARstart{M}{ulti-segment} suction-based soft robots combine geometric adaptability, compliance,
and safe contact, making them attractive for inspection and maintenance in confined or fragile environments
such as aircraft cavities, pipelines, and other cluttered structures~\cite{sunSoftMobileRobots2021,quRecentAdvancesUnderwater2024}.
Compared with rigid or modular systems, they can conform to irregular surfaces,
distribute contact loads, and maintain adhesion while traversing walls, ceilings, holes, and transitions among them~\cite{wanDesignAnalysisRealtime2023,meiSoftCrawlingRobot2024}.
Autonomous locomotion on such surfaces remains difficult.
Existing systems still rely largely on hand-designed gaits or pre-scripted motion sequences
tied to specific setups~\cite{evenLocomotionObstacleAvoidance2023,ketchumAutomatedGaitGeneration2023}.
For suction-based multi-segment soft robots, discrete adhesion transitions and continuous segment deformations are tightly coupled:
small changes in segment curvature or elongation can invalidate a feasible contact transition.
The planning problem is challenging because it must jointly reason about high-dimensional body deformation, discrete adhesion switching,
and online adaptation under uncertainty during physical execution~\cite{calistiFundamentalsSoftRobot2017,shahSoftRobotThat2020,marquezHardwareintheloopSoftRobotic2023}.

\begin{figure}[tp!]
  \centering
  \includegraphics[width=0.86\linewidth]{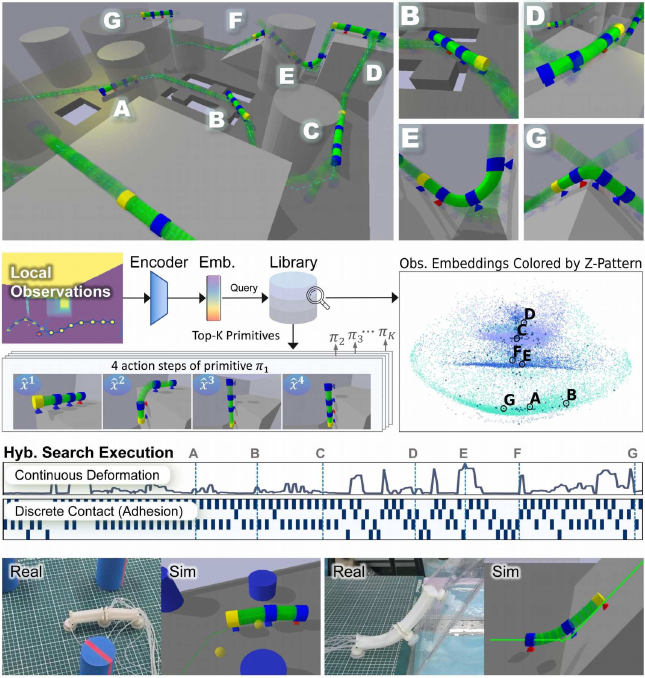}
  \vspace{-1mm}
  \caption{PaHS plans over discrete adhesion modes and continuous body deformation for multi-segment soft robots on diverse 3D terrain transitions.}
  \label{fig:intro}
\end{figure}

\IEEEpubidadjcol

\subsection{Related Work}\label{subsec:intro-related}

(I) \emph{Motion planning via gait generation.}
A large body of work on soft-robot locomotion focuses on generating motion through
deformation patterns, predefined gaits, or morphology-specific controllers~\cite{sunSoftMobileRobots2021,calistiFundamentalsSoftRobot2017}.
Representative approaches tune gait parameters by offline calibration or search~\cite{meiSoftCrawlingRobot2024},
execute hand-crafted sequences through discrete switching logic~\cite{evenLocomotionObstacleAvoidance2023},
or optimize bio-inspired oscillatory controllers with learning-based methods~\cite{xinReinforcementLearningBased2022}.
These methods have shown effective locomotion on particular platforms and terrains, but
they largely treat planning as gait selection or tuning, rather than explicit long-horizon
contact--motion planning under changing 3D geometry and contact constraints.

(II) \emph{Explicit modeling and optimization.}
Another line of work incorporates explicit robot models into planning. Reduced-order
kinematic approximations, such as constant-curvature or multi-rigid-body models, make
search-based motion planning tractable~\cite{luoMotionPlanningIterative2020}, while
higher-fidelity optimization with continuum or finite-element models improves physical
consistency at substantially higher computational cost~\cite{bernTrajectoryOpt2019}.
Although these approaches improve modeling accuracy, they are often expensive and
morphology-dependent, which limits their suitability for long-horizon planning problems
that require repeated feasibility checks during online search.

(III) \emph{Hybrid contact and motion reasoning.}
Compared with gait generation or pure motion synthesis, much less work explicitly addresses
the hybrid nature of suction-based locomotion, where one must jointly decide
\emph{where and when} contacts are established and \emph{how} the body deforms between them.
Rigid multi-contact planners likewise account for contact-dependent
kinematics and stability, either by constructing whole-body contacts along a
planned root trajectory, as for ANYmal, or by jointly optimizing contacts,
gait, and body motion~\cite{geisertContactPlanning2019,
aceitunoSimultaneousContact2019}. Here, each suction mode repartitions a serial
deformable backbone, changing the boundary and load-feasibility conditions of
its continuous deformation problem.
Adhesion-based robots such as MultiTrack and Limpet-II demonstrate the value of
suction-assisted multi-segment mobility~\cite{leeMultiTrack2015,sayedLimpetII2021},
but their planners are often tied to specific mechanisms and transition patterns.
This challenge is amplified in pneumatic soft robots, where delayed nonlinear
deformation and surface-dependent suction introduce substantial uncertainty~\cite{robertsonVSPA2017,liuUnderActuatedAdhesion2022}.
Thus, long-horizon 3D surface-constrained navigation remains underexplored for
suction-based multi-segment soft robots that must jointly reason over adhesion
switching and body deformation.

\subsection{Our Method}
This work proposes a hierarchical hybrid planner for suction-based soft robots
on complex 3D surfaces. The key idea is to combine global near-surface guidance
with contact-aware local planning, so that long-horizon navigation is directed
by surface progress while adhesion, deformation, collision, and suction
feasibility are enforced locally. The core block-wise IK hybrid search (IKHS)
exploits anchors induced by adhesion modes to reduce whole-body hybrid planning
to localized IK over free body blocks. Primitive-augmented hybrid search (PaHS)
accelerates IKHS by retrieving reusable local motion patterns from validated
rollouts, while IK refinement and IKHS fallback preserve feasibility and search
coverage. The resulting framework combines structured hybrid search,
data-driven proposals, and physics-aware validation for simulation and hardware
surface-navigation tasks.

The contributions are twofold: (I) IKHS, a block-wise hybrid planner
for long-horizon contact--motion planning of suction-based multi-segment
soft robots on complex 3D surfaces; and (II) PaHS, which augments IKHS with a
primitive library, learned retrieval, IK refinement, and fallback branching to
improve planning efficiency on recurring local maneuvers.

\section{Problem Description}\label{sec:problem}

\subsection{Robot State, Contacts, and Kinematics}\label{subsec:state-contacts-kin}

\subsubsection{Hybrid state}
\begin{figure}[tp!]
  \centering
  \includegraphics[width=0.95\linewidth]{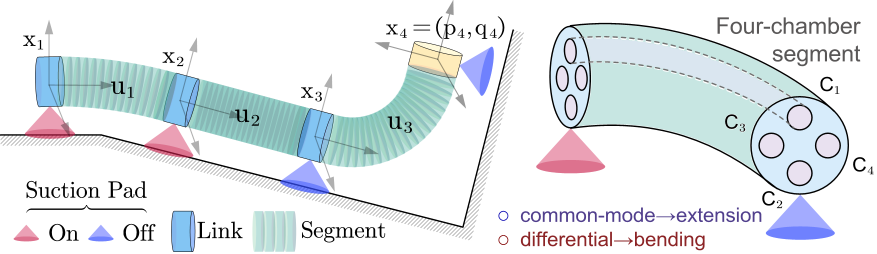}
  \vspace{-0.25mm}
  \caption{
  Multi-segment soft robot with rigid links, four-chamber deformable
  segments, and independently switched suction pads, composed into the hybrid
  state $(\mathbf{X}^h,\mathbf{U}^h,\mathbf{Z}^h)$. Links are indexed from tail to head.
  }
  \vspace{-0.5mm}
  \label{fig:structure}
\end{figure}

As shown in Fig.~\ref{fig:structure}, the robot has~$n_{\texttt{l}}\ge2$
rigid links and~$n_{\texttt{s}}\triangleq n_{\texttt{l}}-1$ deformable
segments. Link~$i$ has pose~$x_i\triangleq(p_i,q_i)\in\mathbb{R}^3\times
\mathbb{S}^3$ and adhesion state~$z_i\in\{0,1\}$, where~$z_i=1$ means
the suction pad is attached. Segment~$i$ is parameterized by a body
twist~$u_i\triangleq(v_i,\omega_i)\in\mathbb{R}^6$, where~$v_i,\omega_i$
are translational and angular terms. At step~$h$, the hybrid robot state is
$(\mathbf{X}^h,\mathbf{U}^h,\mathbf{Z}^h)$ with
$\mathbf{X}^h \triangleq (x_1^h,\cdots,x_{n_{\texttt{l}}}^h)$,
$\mathbf{U}^h \triangleq (u_1^h,\cdots,u_{n_{\texttt{s}}}^h)$, and
$\mathbf{Z}^h \triangleq (z_1^h,\cdots,z_{n_{\texttt{l}}}^h)$.

\subsubsection{Surface and adhesion constraints}
The environment is represented by a signed distance field (SDF)
$D:\mathbb{R}^3\rightarrow\mathbb{R}$ with zero-level surface
$\mathcal{S}\triangleq\{p\in\mathbb{R}^3\mid D(p)=0\}$. Each suction pad has fixed
body offset~$r_i$ and body-frame normal~$\bar n_i$, and its world-frame position and
normal are given by:
\begin{equation*}
\widehat p_i^h\triangleq p_i^h+R(q_i^h)r_i,
\qquad
\widehat n_i^h\triangleq R(q_i^h)\bar n_i,
\end{equation*}
where~$R(q_i^h)$ is the rotation induced by the link orientation~$q_i^h$,
$\widehat p_i^h$ is the world-frame pad position, and~$\widehat n_i^h$ is the
corresponding world-frame pad normal. For an attached pad, the current pose must satisfy
surface proximity and normal alignment, i.e.,
\begin{equation}\label{eq:adhesion-geom}
(z_i^h=1) \;\Rightarrow\;
\begin{cases}
D(\widehat p_i^h)\le\epsilon_{\texttt{d}},\\[1mm]
\|\widehat n_i^h\times n(\widehat p_i^h)\|\le\epsilon_{\texttt{a}},
\end{cases}
\qquad \forall h,i;
\end{equation}
where~$n(\widehat p_i^h)$ is the SDF normal evaluated at~$\widehat p_i^h$,
$\epsilon_{\texttt{d}}$ is the distance threshold, and
$\epsilon_{\texttt{a}}$ is the normal-alignment threshold. If a pad remains attached
across two consecutive steps, its world pose should vary only slightly, i.e.,
\begin{equation}\label{eq:adhesion-stick}
(z_i^h=z_i^{h+1}=1) \;\Rightarrow\;
(\|\widehat x_i^{h+1}-\widehat x_i^h\|\le\epsilon_{\texttt{f}}),
\qquad \forall h,i;
\end{equation}
where~$\widehat x_i^h$ denotes the world-frame pose representation of pad~$i$; and~$\epsilon_{\texttt{f}}$ bounds the allowable inter-step pose variation
under persistent adhesion.
Besides adhesion feasibility, the robot must remain collision-free with respect to the environment,
except for intended suction contact, i.e.,
\begin{equation}\label{eq:collision-free}
D(p)\ge \epsilon_{\texttt{c}}, \quad
\forall p \in \mathcal{R}(\mathbf{X}^h,\mathbf{U}^h)\setminus \mathcal{P}(\mathbf{Z}^h),\ \forall h;
\end{equation}
where $\mathcal{R}(\mathbf{X}^h,\mathbf{U}^h)$ denotes the occupied robot geometry
and~$\mathcal{P}(\mathbf{Z}^h)$ denotes the suction-contact patches.

\subsubsection{Segment kinematics via screw theory}
Let~$T_i\in\mathsf{SE}(3)$ be the homogeneous transform of link~$i$. Each segment is
modeled by the matrix exponential below:
\begin{equation*}
T_{i+1}=T_i\,\Phi(u_i)
=T_i\,\textsf{exp}\!\Bigg(
\begin{bmatrix}
[\omega_i]_\times & v_i\\
0 & 0
\end{bmatrix}
\Bigg),
\end{equation*}
where~$u_i\triangleq(v_i,\omega_i)\in\mathbb{R}^6$ is the body twist of
segment~$i$. Conversely, the adjacent-link twist is recovered by:
\begin{equation}\label{eq:screw}
x_{i+1}^h\triangleq \textsf{Scr}(x_i^h,u_i^h),
\qquad
u_i^h\triangleq \textsf{ScrInv}(x_{i+1}^h,x_i^h);
\end{equation}
where~$\textsf{Scr}(\cdot)$ and~$\textsf{ScrInv}(\cdot)$ denote screw-theoretic
forward propagation and inverse recovery, respectively. The body-fixed local
$z$-axis follows the robot tangent, while the local $x$-axis aligns with the
suction axis.

For the pneumatic platform, tangential extension and bending about the
transverse axes dominate with negligible torsion.
Thus, the reduced prior is given by:
\begin{equation}\label{eq:twist-prior}
u_i\approx[0,\,0,\,v_{z,i},\,\omega_{x,i},\,\omega_{y,i},\,0]^\top,
\end{equation}
by which $L_i\triangleq v_{z,i}$ and
$\boldsymbol{\kappa}_i\triangleq
[\omega_{x,i},\omega_{y,i}]^\top/L_i$ define the segment length and curvature.
Thus, $L_i\|\boldsymbol{\kappa}_i\|$ yields the bending angle of the circular-arc
approximation. For the four-chamber prototype, this reduced prior matches the
dominant actuation modes, while other suction robots may have different priors.

\subsubsection{Load--deformation feasibility}

For a segment deformation to be physically realizable, a gravity-conditioned
quasi-static feasibility constraint is enforced by:
\begin{equation}\label{eq:load-def}
(\overline M_i^h,\,L_i^h,\,\boldsymbol{\kappa}_i^h)
\in \mathcal{S}_{\texttt{D}}(\gamma_i^h),
\end{equation}
where $L_i^h$ and $\boldsymbol{\kappa}_i^h$ are induced by
\eqref{eq:twist-prior}, $\overline M_i^h$ is the equivalent load moment, and
$\gamma_i^h$ is the gravity angle in the fixed-end frame. Since the dominant
free-chain load is gravity, the force direction is determined by
$\gamma_i^h$ and is represented through $\overline M_i^h$, rather than treated
as an independent coordinate.
The family $\mathcal{S}_{\texttt{D}}(\cdot)$ is identified offline
with on-hardware measurements, as described in the sequel.

\subsection{Long-horizon Hybrid Optimization Problem}\label{subsec:objective}

Let the system start from state~$(\mathbf{X}^0,\mathbf{U}^0,\mathbf{Z}^0)$.
The complete sequence~$({\mathbf{X}}^{1:H},{\mathbf{U}}^{1:H},{\mathbf{Z}}^{1:H})$ is optimized to reach the goal
region, and minimize deformation and unsafe contact, i.e.,
\begin{equation}\label{eq:objective}
\begin{aligned}
\underset{{\mathbf{X}}^{1:H},{\mathbf{U}}^{1:H},{\mathbf{Z}}^{1:H}}{\textbf{min}}\quad
& \sum_{h=1}^{H}\!\left(
\mathcal{L}_{\texttt{d}}(\mathbf{U}^{h})+
\mathcal{L}_{\texttt{c}}(\mathbf{X}^{h},\mathbf{Z}^{h})
\right) \\
\text{s.t.}\quad
& \|p_{\texttt{head}}(\mathbf{X}^{H})-p_{\texttt{goal}}\|\le r_{\texttt{g}},\\
& \eqref{eq:adhesion-geom},\eqref{eq:adhesion-stick},\eqref{eq:collision-free},\eqref{eq:screw},\eqref{eq:load-def};
\end{aligned}
\end{equation}
where~$H$ is the planning horizon;
$\mathcal{L}_{\texttt{d}}(\mathbf{U}^{h})$ regularizes deformation effort
by favoring lower-curvature and lower-extension solutions among contact
transitions with similar guide progress;
$\mathcal{L}_{\texttt{c}}(\mathbf{X}^{h},\mathbf{Z}^{h})$ penalizes unsafe
contact; $p_{\texttt{head}}(\mathbf{X}^{H})$ is the terminal head position; and
$(p_{\texttt{goal}},r_{\texttt{g}})$ define the goal region.
This formulation
captures long-horizon hybrid planning for both contact and motion, under the
coupled adhesion, kinematic, and load--deformation constraints.

\begin{remark}\label{rm:problem}
At each step, the planner must choose both a binary contact
$\mathbf{Z}^h\in\{0,1\}^{n_{\texttt{l}}}$ and a continuous deformation $\mathbf{U}^h\in\mathbb{R}^{6 n_{\texttt{s}}}$. Over a horizon $H$,
this yields a hybrid decision space
$\big(\{0,1\}^{n_{\texttt{l}}}\times\mathbb{R}^{6 n_{\texttt{s}}}\big)^H$.
In addition, the collision, adhesion, and load
constraints are nonconvex and coupled across steps, so local feasibility
depends on future configurations. Thus, exact planning for
\eqref{eq:objective} is intractable in general.
\hfill $\blacksquare$
\end{remark}

\section{Proposed Solution}\label{sec:solution}
The proposed framework has three components. Sec.~\ref{subsec:ikhs}
introduces IKHS, which exploits the anchor structure induced by adhesion
to reduce full-chain hybrid
expansion to validated block-wise updates. Sec.~\ref{subsec:learn-prim} then explains
how short validated motion primitives are stored and reused in PaHS as a
retrieval-based acceleration for IKHS. The overall
analyses including online execution are provided in Sec.~\ref{subsec:overall}.

\subsection{Core Planner: Block-wise IK Hybrid Search (IKHS)}
\label{subsec:ikhs}

This subsection presents IKHS, the core hybrid planner in the proposed
framework. It uses a guiding surface path for progress and local target proposals,
while exploiting the free-block structure induced by the current adhesion mode.
The resulting best-first procedure, summarized in Alg.~\ref{alg:hybrid_planner}
and Fig.~\ref{fig:bw_ikhs}, also serves as the fallback expansion in PaHS.

\subsubsection{Guiding surface path}\label{subsubsec:guide-path}
\begin{figure}[t!]
  \centering
  \includegraphics[width=1.0\linewidth]{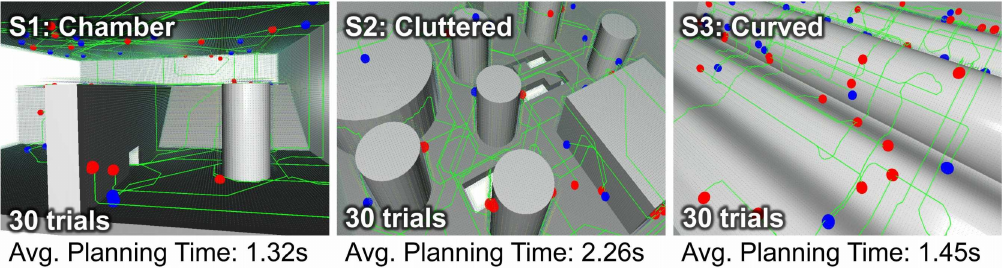}
  \vspace{-2mm}
  \caption{Examples of the guiding surface paths in three test scenarios.}
  \label{fig:guiding_path}
  \vspace{-0.25mm}
\end{figure}

The guide is computed on an augmented SDF-derived graph
$\bar{\mathcal{G}}_{\texttt{s}}$, whose locations $\hat p$ are lattice points in the
near-surface band
$\{\hat p\in\mathbb{R}^3 \mid d_{\min}\le D(\hat p)\le d_{\max}\}$, with
$d_{\min}=0.005\,\mathrm{m}$ and $d_{\max}=0.025\,\mathrm{m}$, allowing small gaps
rather than restricting search to the zero-level surface $\mathcal{S}$.
Since some guide penalties depend on local approach geometry, the same
near-surface location may incur different downstream costs when reached from
different directions. Each node $\bar v=(\hat p,\delta)$ therefore includes a
discretized incoming direction, while a short ancestor history distinguishes
penalties on recent tangents, turns, and surface switches. From
$\bar v_{\texttt{s}}=(\hat p_0,\delta_0)$, where
$\hat p_0\triangleq p_{\texttt{head}}(\mathbf{X}^0)$ and $\delta_0$ is the initial
heading, search proceeds to $p_{\texttt{goal}}$ and returns
$\mathcal{P}\triangleq(\hat p_0,\cdots,\hat p_T)$ for progress measurement and
local target proposals.
The path is generated by~A$^\star$ search with the edge cost defined below:
\begin{equation}\label{eq:guide-cost}
\begin{aligned}
c_{t\rightarrow t+1}\triangleq&\;w_\ell d_{t\rightarrow t+1}
+w_{\texttt{lift}}\phi_{\texttt{lift}}
+w_{\texttt{sharp}}\phi_{\texttt{sharp}} \\
&+w_{\texttt{n}}\phi_{\texttt{normal}}
+w_{\texttt{b}}\phi_{\texttt{bend}}
+w_{\texttt{nc}}\phi_{\texttt{nc}},
\end{aligned}
\end{equation}
where $d_{t\to t+1}$ is the edge length, $\phi_{\text{lift}}$ penalizes loss of
surface support, $\phi_{\text{sharp}}$ measures local surface irregularity,
$\phi_{\text{normal}}$ penalizes surface-normal change,
$\phi_{\text{bend}}$ is the in-plane turning angle over a short history,
and $\phi_{\text{nc}}$ penalizes non-coplanar transitions between
path direction and adjacent surface normals. These terms bias the guide
toward contact-compatible routes near ridges, wall--ceiling transitions,
and obstacle boundaries, as shown in Fig.~\ref{fig:guiding_path}.
The guide planner is implemented in C++ on a 26-neighbor voxel graph.



\subsubsection{Adhesion-induced block decomposition}
\label{subsubsec:kinematic}
\begin{figure}[t!]
  \centering
  \includegraphics[width=0.97\linewidth]{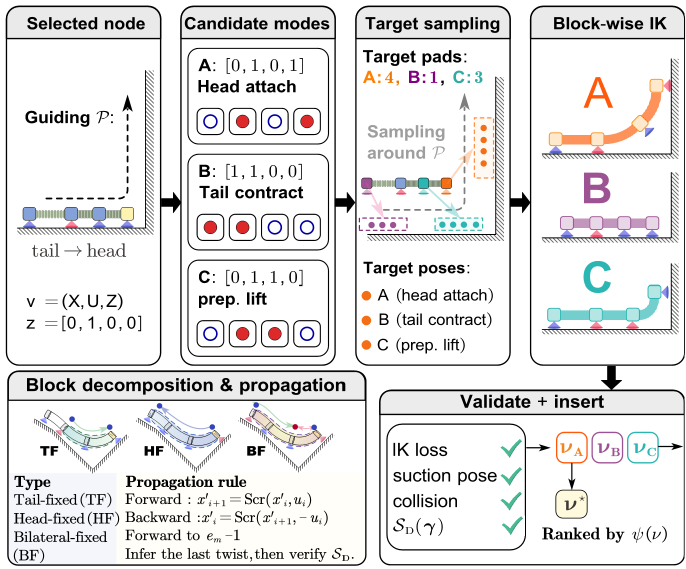}
  \vspace{-1mm}
  \caption{
    Schematic of the block-wise IK hybrid search (IKHS),
      including five main steps (\textbf{top});
      the block decomposition and propagation (\textbf{bottom}).}
  \label{fig:bw_ikhs}
  \vspace{-1mm}
\end{figure}

Given the adhesion mode~$\mathbf{Z}$, the chain is partitioned into consecutive blocks:
\begin{equation}\label{eq:block-decomp}
\big\{\mathcal{B}_m\big\}_{m=1}^{M}
\triangleq \textsf{BlockDecomp}(\mathbf{Z}),
\end{equation}
where $\mathcal{B}_m\triangleq [s_m,\, e_m]$ and attached links serve as kinematic anchors.
As summarized in Fig.~\ref{fig:bw_ikhs},
a free block may be head-fixed, tail-fixed, or bilateral-fixed,
and each type follows a different propagation rule under the screw model in~\eqref{eq:screw}.
Thus, given the current chain pose~$\mathbf{X}$ and a candidate segment-twist vector~$\mathbf{U}'$,
the corresponding block update is denoted by:
\begin{equation}\label{eq:block-update}
\mathbf{X}'_m \triangleq \textsf{BlockProp}(\mathbf{X},\, \mathbf{U}',\, \mathcal{B}_m),
\quad m=1,\cdots,M;
\end{equation}
The updated block states are merged into the full chain pose~$\mathbf{X}'$
given the attached links. This anchor-induced decomposition is the key planning
reduction: it converts full-chain hybrid expansion into local block updates while
preserving downstream feasibility.

\subsubsection{Identify Load-deformation Feasibility Manifold~$\mathcal{S}_{\texttt{D}}(\cdot)$}
\label{subsubsec:S-D}
The load-deformation feasibility manifold~$\mathcal{S}_{\texttt{D}}(\cdot)$
in~\eqref{eq:load-def} is essential for the hybrid search and is identified offline.
Approximately $10^3$ hardware measurements are used to calibrate a finite-element
segment model, which then generates about $5\times10^5$ quasi-static samples.
The gravity angle is discretized into $15^\circ$ intervals, and approximately
$6\times10^4$ uncertain near-boundary samples are discarded before fitting one
conservative convex polytope in $(\overline M,L,\boldsymbol{\kappa})$ for each
interval. During planning, each segment selects the polytope corresponding to its
current gravity angle. The same family is used for all nominally identical
four-chamber segments, while multi-segment coupling is checked through full-chain
propagation, closure, adhesion, and collision constraints.

\subsubsection{Block-wise IK hybrid search (IKHS)}
\label{subsubsec:ikhs}

IKHS performs best-first search over hybrid states
$\{\nu_h=(\mathbf{X}^h,\mathbf{U}^h,\mathbf{Z}^h)\}$. Given start node~$\nu_0$,
goal~$p_{\texttt{g}}$, and guide~$\mathcal{P}$, it backtracks the best goal-reaching
node to obtain
$\boldsymbol{\tau}^\star=(\mathbf{X}^{1:H},\mathbf{U}^{1:H},\mathbf{Z}^{1:H})$
(Alg.~\ref{alg:hybrid_planner}). Frontier nodes~$\mathcal{F}$ are scored by:
\begin{equation}\label{eq:ikhs-select}
\psi(\nu_h)\triangleq
\mathcal{J}_{\texttt{prg}}(\nu_h)-w_{\texttt{d}}^h\mathcal{J}_{\texttt{def}}(\nu_h)
+w_{\texttt{e}}\mathcal{J}_{\texttt{exp}}(\mathbf{Z}^h),
\end{equation}
where~$\mathcal{J}_{\texttt{prg}}\triangleq
\Delta s_{\mathcal{P}}-w_{\texttt{lat}}\Delta d_{\mathcal{P}}$;
$\Delta s_{\mathcal{P}}$ and $\Delta d_{\mathcal{P}}$ denote the cumulative
changes from the start node to~$\nu_h$ in link arc-length progress along~$\mathcal{P}$
and lateral guide deviation, respectively;
$\mathcal{J}_{\texttt{def}}$ accumulates normalized extension and squared
bending, with adaptive weight~$w_{\texttt{d}}^h$ reduced when recent progress
stalls; and~$\mathcal{J}_{\texttt{exp}}$ is a bounded sibling-level bonus for
under-expanded adhesion modes. Numerical values are reported in
Table~\ref{tab:impl-params}.
Thus the node within~$\mathcal{F}$ is selected by:
\begin{equation}\label{eq:ikhs-argmax}
\nu \triangleq \textbf{argmax}_{\tilde{\nu}\in\mathcal{F}}\, \big{\{}\psi(\tilde{\nu})\big{\}},
\end{equation}
For a selected node~$\nu=(\mathbf{X},\mathbf{U},\mathbf{Z})$,
the current adhesion mode first induces a free-block decomposition
$
\{\mathcal{B}_m\}_{m=1}^{M}\triangleq \textsf{BlockDecomp}(\mathbf{Z}),
$
which defines the local structure for expansion.
Function~$\textsf{ZCand}(\nu)$ then enumerates the admissible successor adhesion
modes~$\mathcal{Z}_{\texttt{cand}}$.
Each candidate mode~$\mathbf{Z}_{\texttt{new}}\in\mathcal{Z}_{\texttt{cand}}$
remains feasible for the current pose and introduces at most one new
attachment within each free block.
Given~$\mathbf{Z}_{\texttt{new}}$ and~$\{\mathcal{B}_m\}_{m=1}^{M}$,
the IK-based expansion generates children by:
\begin{equation}\label{eq:ikexpand-op}
\mathcal{V}_{\texttt{ik}}
\triangleq
\textsf{IKExpand}\big{(}\nu,\,\mathbf{Z}_{\texttt{new}},\,\{\mathcal{B}_m\}_{m=1}^{M},\,\mathcal{P}\big{)}.
\end{equation}
where~$\textsf{IKExpand}(\cdot)$ operates block-wise:
for each affected free block~$\mathcal{B}_m$, it samples target poses for the
designated new attachment pad from a short forward window on the guiding
path~$\mathcal{P}$ and solves reduced IK only for that block.
Specifically, the block deformation variables~$\mathbf{U}'_m$ are obtained by
\begin{equation}\label{eq:perblock-ik}
\begin{aligned}
\underset{\mathbf{U}'_m}{\textbf{min}}\;&
\mathcal{L}_{\texttt{tar}}(\mathbf{x}'_{j^\star},\mathbf{x}^\star)
+w_d\,\mathcal{L}_{\texttt{def}}(\mathbf{U}'_m) \\
\textbf{s.t.}\;&
\mathbf{X}'_m=\textsf{BlockProp}(\mathbf{X},\mathbf{U}'_m,\mathcal{B}_m), \\
&
(\overline M_i',\,L_i',\,\boldsymbol{\kappa}_i')
\in \mathcal{S}_{\texttt{D}}(\gamma_i'),\quad \forall i\in \mathcal{B}_m ,
\end{aligned}
\end{equation}
where $\mathbf{x}^\star$ is the sampled target pose and
$\mathbf{x}'_{j^\star}$ is the resulting pose of the designated new
attachment pad;
$\mathcal{L}_{\texttt{tar}}$ is the sum of four weighted error terms:
the normal-distance term
$w_{\mathrm{nd}}\mathcal{L}_{\mathrm{nd}}$,
the tangential-distance term
$w_{\mathrm{td}}\mathcal{L}_{\mathrm{td}}$,
the surface-normal orientation term
$w_{\mathrm{no}}\mathcal{L}_{\mathrm{no}}$,
and the full-orientation term
$w_{\mathrm{o}}\mathcal{L}_{\mathrm{o}}$;
and the deformation loss~$\mathcal{L}_{\texttt{def}}$ regularizes excessive
segment deformation.
The reduced IK variables follow~\eqref{eq:twist-prior}.
Condition~\eqref{eq:perblock-ik} is implemented using a
load-conditioned convex section of
$\mathcal{S}_{\texttt{D}}(\gamma)$ at the parent state.
The fused child is accepted only after passing the IK-loss, suction-pose,
collision, and load--deformation checks, with the relevant
quantities recomputed at the child state. Implementation parameters and
weights are summarized in Table~\ref{tab:impl-params}.

\begin{figure*}[t!]
  \centering
  \includegraphics[width=0.95\hsize]{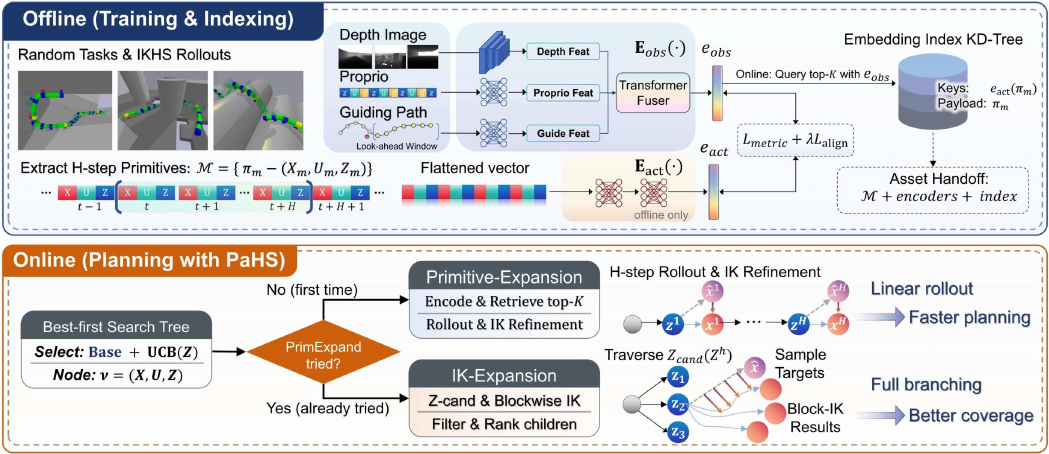}
  \vspace{-0.25mm}
  \caption{Unified overview of IKHS and PaHS. Offline IKHS rollouts generate
  validated short-horizon primitives. Online, IKHS performs best-first
  hybrid search with block-wise IK, while PaHS first attempts
  primitive-based expansion and falls back to IKHS when needed.}
  \label{fig:overall}
\end{figure*}

\begin{algorithm}[t!]
  \caption{Hybrid planner for IKHS and PaHS}
  \label{alg:hybrid_planner}
  \SetAlgoLined
  \KwIn{Start node~$\nu_0$, goal~$p_{\texttt{g}}$, guiding path~$\mathcal{P}$,
  flag~\texttt{useRet}, and library~$\mathcal{M}$ (optional).}
  \KwOut{Plan~$\boldsymbol{\tau}^\star=(\mathbf{X}^{1:H},\mathbf{U}^{1:H},\mathbf{Z}^{1:H})$.}
  Initialize frontier
  $\mathcal{F}\leftarrow\{\nu_0\}$\;
  Set $\textsf{PrimExpanded}(\nu_0)\leftarrow \texttt{False}$\;
  \While{$\mathcal{F}\neq\emptyset$ and budget not exhausted}{
    $\nu\leftarrow \arg\max_{\tilde{\nu}\in\mathcal{F}}\,\psi(\tilde{\nu})$\;
    \If{$\textsf{GoalReached}(\nu,\,p_{\texttt{g}})$}{\textbf{break}}
    \If{$\texttt{useRet}$ \textbf{and} $(\neg\textsf{PrimExpanded}(\nu)$)}{
      $\mathcal{V}_{\texttt{prim}}\leftarrow
      \textsf{PrimExpand}(\nu,\,\mathcal{M},\,\mathcal{P})$\;
      Insert valid children~$\mathcal{V}_{\texttt{prim}}$ into~$\mathcal{F}$\;
      $\textsf{PrimExpanded}(\nu)\leftarrow \texttt{True}$\;
      \textbf{Continue}\;
    }
    $\mathcal{Z}_{\texttt{cand}}\leftarrow \textsf{ZCand}(\nu)$\;
    $\{\mathcal{B}_m\}_{m=1}^{M}\leftarrow \textsf{BlockDecomp}(\mathbf{Z}^{\nu})$\;
    \ForEach{$\mathbf{Z}_{\texttt{new}}\in\mathcal{Z}_{\texttt{cand}}$}{
      $\mathcal{V}_{\texttt{ik}}\leftarrow
      \textsf{IKExpand}(\nu,\,\mathbf{Z}_{\texttt{new}},\,\{\mathcal{B}_m\}_{m=1}^{M},\,\mathcal{P})$\;
      Insert valid children~$\mathcal{V}_{\texttt{ik}}$ into~$\mathcal{F}$\;
    }
    Remove~$\nu$ from~$\mathcal{F}$\;
  }
  Backtrack to compute~$\boldsymbol{\tau}^\star$\;
\end{algorithm}

Finally, the search terminates when the frontier becomes empty, or when the
iteration budget is exhausted, or when
function~$\textsf{GoalReached}(\nu,\,p_{\texttt{g}})$ returns true, i.e., when the
current state satisfies the prescribed goal tolerance with respect to the goal
position~$p_{\texttt{g}}$. Under finite search budget, beam width, and
per-block candidate bounds, IKHS terminates in finite time and returns only
plans that satisfy the modeled kinematic, adhesion, collision
and load--deformation constraints.

\subsection{Primitive-augmented hybrid search (PaHS)}
\label{subsec:learn-prim}

As summarized in Fig.~\ref{fig:overall}, PaHS retains the IKHS search tree,
node score~\eqref{eq:ikhs-select}, guide~$\mathcal{P}$, and stopping rule, while
augmenting node expansion with short feasible contact--motion continuations. The
key idea is to reuse local motion patterns already validated by IKHS while
preserving the same downstream feasibility checks.
\begin{figure}[t!]
  \centering
  \includegraphics[width=0.97\linewidth]{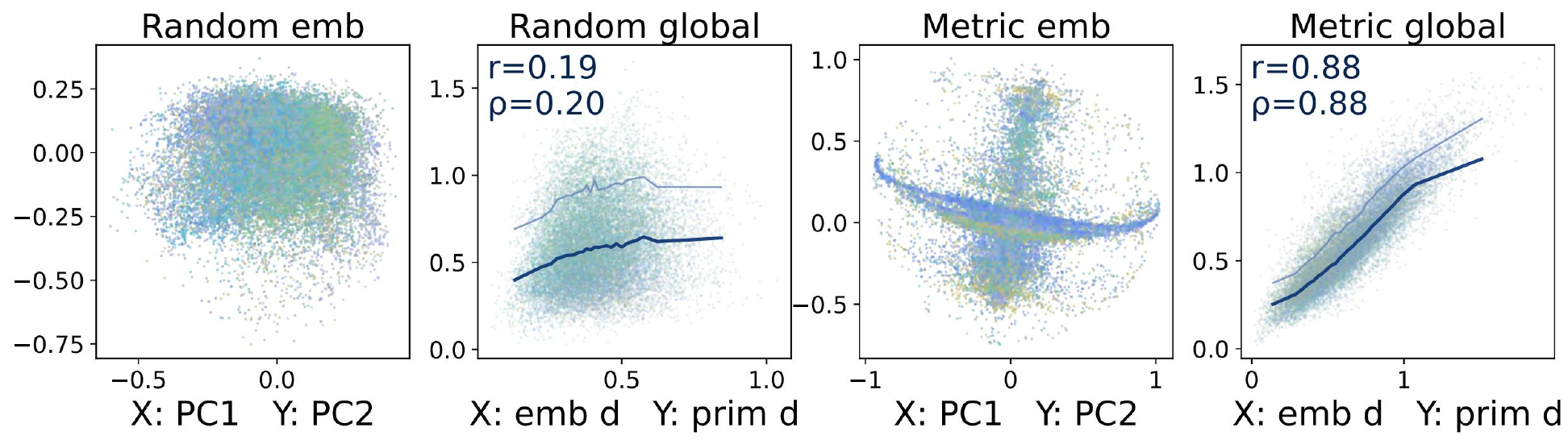}
  \vspace{-0.5mm}
  \caption{Different embeddings for primitives: random (left) and learned
  (right), compared by their alignment with primitive distances.}
  \vspace{-0.5mm}
  \label{fig:metric_vis}
\end{figure}

\subsubsection{Primitive library}
The primitive library is constructed offline from successful IKHS rollouts.
As shown in Fig.~\ref{fig:overall}, given a rollout
$\boldsymbol{\tau}^\star$,
the fixed-horizon segments are extracted as short contact--motion primitives, i.e.,
\[
\pi_h \triangleq
(\mathbf{X}^{h:h+H_{\texttt{p}}},
\mathbf{U}^{h:h+H_{\texttt{p}}-1},
\mathbf{Z}^{h:h+H_{\texttt{p}}}),
\quad
0\le h\le H-H_{\texttt{p}};
\]
where $H_{\texttt{p}}$ is the primitive horizon. Each primitive therefore
records a verified local continuation of body deformation and adhesion switching
over a short horizon. It is expressed in the head frame of its first step,
suppressing absolute pose variation so retrieval depends mainly on the local
contact--motion pattern. The resulting library is:
\begin{equation}\label{eq:memory}
\mathcal{M}\triangleq \big{\{}(\mathbf{e}^j_{\texttt{act}},\,\pi_j)\big{\}}_{j=1}^{N_{\mathcal{M}}},
\end{equation}
where $\pi_j$ is a stored primitive;
$\mathbf{e}^j_{\texttt{act}}\in\mathbb{R}^d$ is its retrieval key; and
$N_{\mathcal{M}}$ is the number of stored primitives.
Each sample contains a head-frame observation, including a local depth image,
proprioception, adhesion states, gravity directions, and a guide look-ahead
window. The associated primitive stores the next $H_{\texttt{p}}=4$ hybrid
states as relative link poses, segment twists, and adhesion modes.

\begin{figure*}[t!]
  \centering
  \includegraphics[width=0.95\hsize]{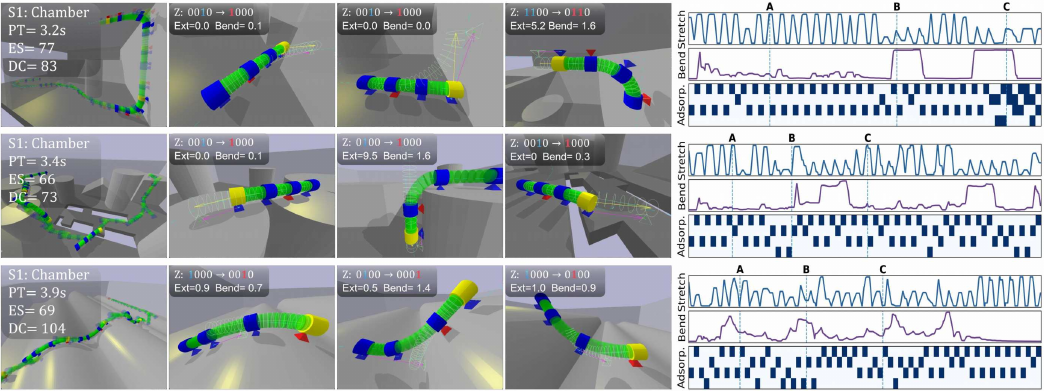}
  \vspace{-0.5mm}
  \caption{Representative simulation scenarios and planned trajectories. The benchmark covers cluttered terrain, strong curvature, and multiple surface transitions.}
  \vspace{-1mm}
  \label{fig:exp_result}
\end{figure*}
\subsubsection{Retrieval embedding}
Retrieval is performed by comparing the current planning context with stored
primitives in a shared latent space. A dual-tower embedding is learned, where
the action tower $E_{\texttt{act}}(\cdot)$ encodes each primitive~$\pi_j\in\mathcal{M}$ from its
canonicalized local rollout descriptor~$\mathbf{r}(\pi)$, and the observation
tower $E_{\texttt{obs}}(\cdot)$ encodes the current search context~$\mathbf{o}(\nu,\mathcal{P})$ from
node~$\nu$ and a short look-ahead window on the guiding path~$\mathcal{P}$.
The two encoders map them into a common normalized $d$-dimensional space:
\[
\mathbf{e}_{\texttt{obs}} \triangleq
\frac{E_{\texttt{obs}}(\mathbf{o}(\nu,\mathcal{P}))}
{\|E_{\texttt{obs}}(\mathbf{o}(\nu,\mathcal{P}))\|_2},
\quad
\mathbf{e}_{\texttt{act}} \triangleq
\frac{E_{\texttt{act}}(\mathbf{r}(\pi))}
{\|E_{\texttt{act}}(\mathbf{r}(\pi))\|_2};
\]
where the $\ell_2$-norm makes similarity comparison scale-invariant;
output~$\mathbf{e}_{\texttt{obs}}$ is the online query used
during planning; and~$\mathbf{e}_{\texttt{act}}$ is the stored retrieval key
in~\eqref{eq:memory}.
The training combines a metric loss~$\mathcal{L}_{\texttt{metric}}$ to
encourage primitives with similar pose sequences to have nearby embeddings, and
an alignment loss~$\mathcal{L}_{\texttt{align}}$ to align each observation
embedding with the primitive corresponding to the same local continuation.
Thus, the training loss is given by:
\[
\mathcal{L}\triangleq
\mathcal{L}_{\texttt{metric}}
+\lambda_{\texttt{align}}\mathcal{L}_{\texttt{align}},
\]
where $\lambda_{\text{align}}>0$ is a weight parameter.

The observation tower combines a $56{\times}56$ depth backbone, MLP
proprioception encoders, and a two-layer four-head transformer fusion
module. The action tower is a two-layer 256-hidden MLP, and both towers output
$\ell_2$-normalized 16-D embeddings. Training uses metric and cosine
alignment losses with hard-pair sampling; rollout episodes are split before
sliding primitive windows are extracted to avoid data leakage.
Fig.~\ref{fig:metric_vis} shows that
the learned embedding aligns latent distances with primitive similarity
better than random features.

\subsubsection{Primitive-augmented expansion}
\label{subsec:sampling}
As shown in Fig.~\ref{fig:overall},
given the learned library~$\mathcal{M}$, for a selected node~$\nu$, PaHS first attempts the primitive-based expansion by:
\begin{equation}\label{eq:primexpand-operator}
\mathcal{V}_{\texttt{prim}} \triangleq
\textsf{PrimExpand}(\nu,\, \mathcal{M},\, \mathcal{P}),
\end{equation}
where the operator encodes $\mathbf{e}_{\texttt{obs}}$, retrieves the top-$K$
nearest primitives from $\mathcal{M}$ by KD-tree, rolls them out for
$H_{\texttt{p}}$ steps, rejects fast geometric or adhesion failures, and refines
survivors by step-wise block IK under the current SDF and guide. If no refined
primitive is valid, \textsf{PrimIKHS} falls back to IKHS. Otherwise, the returned
children are inserted and standard IK expansion is skipped for that iteration.
Primitive expansion is attempted at most once per node; if the node is selected
again later, expansion proceeds with standard IKHS branching. This design keeps
retrieval lightweight while preserving the same collision, adhesion, and
load--deformation validation used in IKHS.

\subsection{Overall Analyses}\label{subsec:overall}

\subsubsection{Online execution and adaptation}
\label{subsubsec:online-exec}
Execution follows the returned plan in a receding-horizon manner. For each
transition, a short deformation schedule $\widetilde{\mathbf{U}}^{h,0:T_h}$ is
constructed under fixed adhesion mode $\mathbf{Z}^h$, and the rollout
$\widetilde{\mathbf{X}}^{h,0:T_h}$ is checked against persistent
adhesion~\eqref{eq:adhesion-stick}, collision safety~\eqref{eq:collision-free},
load--deformation feasibility~\eqref{eq:load-def}, and the same geometric
checks used in IKHS/PaHS. Segment-level tracking uses the clipped integral pressure
commands below:
\begin{equation}\label{eq:int-control}
\Delta \mathbf{p}^{h,t}_i
=
\textsf{clip}\!\left(
K_i^{\texttt{len}} I^{h,t}_{\texttt{len},i}
+
K_i^{\texttt{rot}} I^{h,t}_{\texttt{rot},i}
\right),
\end{equation}
where $I^{h,t}_{\texttt{len},i}$ and $I^{h,t}_{\texttt{rot},i}$ accumulate
extension and bending errors. The switch
$\mathbf{Z}^h\!\rightarrow\!\mathbf{Z}^{h+1}$ is issued only after the terminal
tracking tolerances are met. If interpolation or tracking violates safety
checks, the rollout is rebuilt with tighter margins. If mismatch persists, a
nearby validated primitive is replayed or a short-horizon PaHS subproblem is
solved from this state.

\subsubsection{Correctness and completeness}
\label{subsubsec:correctness}
IKHS is correct with respect to the modeled constraints: every accepted child is
generated by block-wise kinematic propagation and inserted only after adhesion,
geometric, and load--deformation validation. PaHS retains this property because
retrieved candidates are accepted only after the same IK refinement and
feasibility checks, otherwise falling back to IKHS. Thus, any returned plan
satisfies~\eqref{eq:adhesion-geom}--\eqref{eq:collision-free}
and~\eqref{eq:load-def}, although no guarantee is claimed for unmodeled effects.
Completeness is limited by finite contact candidates, sampled guide targets,
bounded IK iterations, beam width, and finite planning budget.

\subsubsection{Overall complexity}\label{subsubsec:complexity}
The A$^\star$ guide computation costs $\mathcal{O}(|E_s|\log |V_s|)$. For IKHS,
per-node runtime is dominated by $\mathcal{Z}_{\texttt{cand}}(\nu)$ enumeration
and block-wise IK over affected free blocks, scaling with candidate modes and
block sizes rather than full-chain IK. PaHS adds $\mathcal{O}(d\log N_M)$
retrieval, top-$K$ rollout/checking over $H_p$ steps, and local refinement,
with fallback to IKHS. Memory is $\mathcal{O}(|\mathfrak{T}|+N_Md)$.

\newcolumntype{Y}{>{\centering\arraybackslash}X}

\section{Numerical Experiments}\label{sec:experiments}

Numerical experiments evaluate planning efficiency, retrieval quality, and
hardware transfer on a laptop with an Intel Core i9-13900HX CPU and NVIDIA RTX
4070 GPU. The Python3 implementation uses \texttt{IPOPT}/\texttt{CasADi} for IK,
\texttt{PyBullet} for visualization and simulated depth, and GPU-accelerated
\texttt{Open3D} for SDF updates. Simulation and hardware videos are available \href{https://youtu.be/OQR5Sx5Bwnc}{online}.

\begin{table}[t]
\centering
\begingroup
\color{black}
\caption{Main implementation parameters.}
\label{tab:impl-params}
\vspace{-1mm}
\scriptsize
\renewcommand{\arraystretch}{1.02}
\setlength{\tabcolsep}{3pt}
\begin{tabularx}{\columnwidth}{
@{}>{\raggedright\arraybackslash}p{0.17\columnwidth}
>{\raggedright\arraybackslash}X@{}
}
\toprule
\textbf{Module} & \textbf{Parameters} \\
\midrule
Guide &
$(w_{\ell},w_{\texttt{lift}},w_{\texttt{sharp}},
w_{\texttt{n}},w_{\texttt{b}},w_{\texttt{nc}})
=(1,1,10,1,2,30)$
\\

IK &
Target weights
$(w_{\mathrm{nd}},w_{\mathrm{td}},w_{\mathrm{no}},w_{\mathrm{o}})
=(10^4,10^2,10^2,10^{-1})$;
deformation weight $w_d=10^{-2}$;
max. $40$ iter./target
\\

Expansion &
$\le 32$ targets/free block;
top-$K$ pruning, $K=3$;
beam $3$
\\

IKHS score &
$w_{\texttt{lat}}=0.4$;
$w_{\texttt{d}}^h=0.1\alpha_{\texttt{p}}$,
$\alpha_{\texttt{p}}\in[0.1,1.0]$;
$w_{\texttt{e}}=0.05$;
ext./bend $(0.2,1.0)$
\\

PaHS &
$H_{\texttt{p}}=4$;
retrieval $K=50$;
$d=16$
\\

Training &
batch $256$;
lr $10^{-3}$;
decay $10^{-5}$;
$200$ epochs;
hard-pair sampling
\\

Feasibility &
$\Delta\gamma=15^\circ$ for
$\mathcal{S}_{\texttt{D}}(\gamma)$;
$(\epsilon_{\texttt{d}},\epsilon_{\texttt{a}})
=(0.5\,\mathrm{cm},\pi/10)$;
$\epsilon_{\texttt{c}}=0.5\,\mathrm{cm}$
\\
\bottomrule
\end{tabularx}
\vspace{-2mm}
\endgroup
\end{table}

\subsection{Simulation Setup}\label{subsec:description}

Planning and execution simulation use a custom transition model
for the robot-environment interaction and contact-state evolution.
Main implementation parameters are summarized in
Table~\ref{tab:impl-params}. Unless otherwise noted, they are fixed across all
simulation tasks.
As shown in Fig.~\ref{fig:exp_result},
the benchmark scenarios have complementary difficulties: \textbf{S1} tests
feasibility under floor-wall-ceiling transitions; \textbf{S2} tests the reuse of recurring local transition patterns in
clutter; and \textbf{S3} tests guide quality and search coverage under large
surface-normal variation. In each scenario, $20$ navigation tasks are evaluated.


\begin{figure}[t!]
  \centering
  \includegraphics[width=0.95\linewidth]{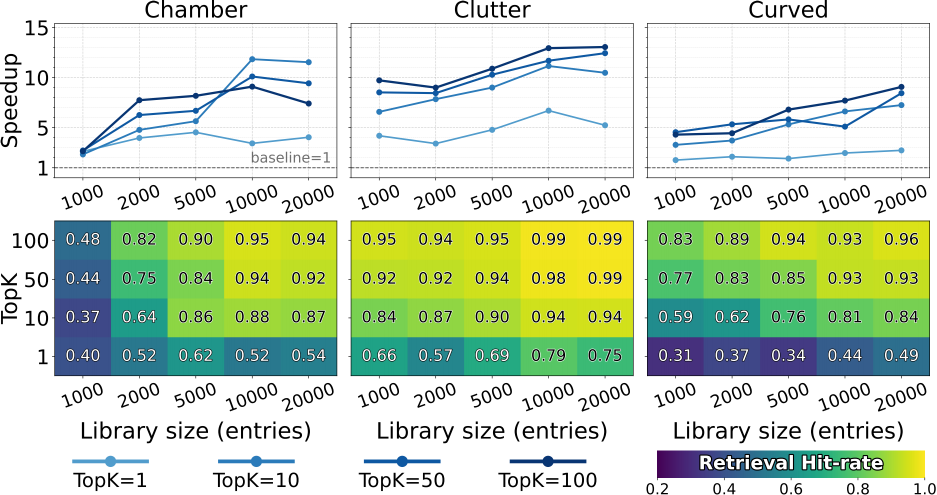}
  \vspace{-0.5mm}
  \caption{Effect of library size and expansion budget on the retrieval hit
  rate and planning acceleration, within three different scenarios.}
  \vspace{-0.5mm}
  \label{fig:sweep_entries_topk}
\end{figure}

\subsection{Results and Analysis}\label{subsec:results}

(I) \textbf{Offline primitive library}: The success-data library for PaHS is
constructed offline from IKHS rollouts in the three benchmark scenarios.
A total of $500$ episodes are collected per scenario, for $1500$ episodes
overall, yielding more than $4\times 10^4$ primitive segments. After embedding,
indexing, and duplicate removal, the final library contains about $6000$
entries. Episodes are split before fixed-horizon windows are extracted
and final evaluation tasks use independent seeds.

(II) \textbf{Overall planning performance}: Across $60$ benchmark tasks, IKHS
and PaHS both achieve 20/20 success in S1--S3, covering clutter, strong curvature,
and multiple surface transitions (Fig.~\ref{fig:exp_result}). The
metrics are gait-planning time (PT), execution steps (ES), and deformation cost
(DC); guide-search times, excluded from PT, are $1.42$, $2.35$, and $1.59\,$s.
PaHS reduces IKHS PT from $20.4$, $11.9$, and $15.6\,$s to $1.7$, $1.0$, and
$1.8\,$s, i.e., $12.0\times$, $11.9\times$, and $8.7\times$ speedups
(Table~\ref{tab:compare_ablation}), with hit rates of $92\%$, $99\%$, and $93\%$.

\begin{table}[t!]
  \caption{Metric-learning ablation for primitive retrieval.}
  \label{tab:metric-ablation}
  \centering
  \footnotesize
  \renewcommand{\arraystretch}{1.0}
  \begin{tabular*}{1\columnwidth}{@{\extracolsep{\fill}}lcc|cc}
    \toprule
    & \multicolumn{2}{c}{Seen (avg. 3 scen.)}
    & \multicolumn{2}{c}{Unseen} \\
    \cmidrule(lr){2-3}\cmidrule(lr){4-5}
    Encoder & PT (s) & Hit (\%) & PT (s) & Hit (\%) \\
    \midrule
    Metric (\textbf{ours}) & 2.64 & 87.9 & 2.88 & 80.5 \\
    Random & 3.25 & 77.5 & 4.09 & 64.3 \\
    Pooled PCA & 5.17 & 62.1 & 4.29 & 58.6 \\
    \bottomrule
  \end{tabular*}
  \vspace{-1mm}
\end{table}

(III) \textbf{Scaling with library size and retrieval budget}:
The retrieval hit rate and planning acceleration improve with library size and
retrieval budget. Under top-$100$ retrieval, increasing the library from
$1000$ to $10{,}000$ entries raises the hit rate from $48\%$ to $95\%$ in S1
and from $83\%$ to $93\%$ in S3, while S2 remains near $97\%$. The gain
saturates beyond about $10^4$ entries or $K=50$. As shown in
Fig.~\ref{fig:sweep_entries_topk}, S3 remains the most difficult but improves
consistently with larger libraries and budgets.

\begin{figure}[t!]
  \centering
  \includegraphics[width=0.92\linewidth]{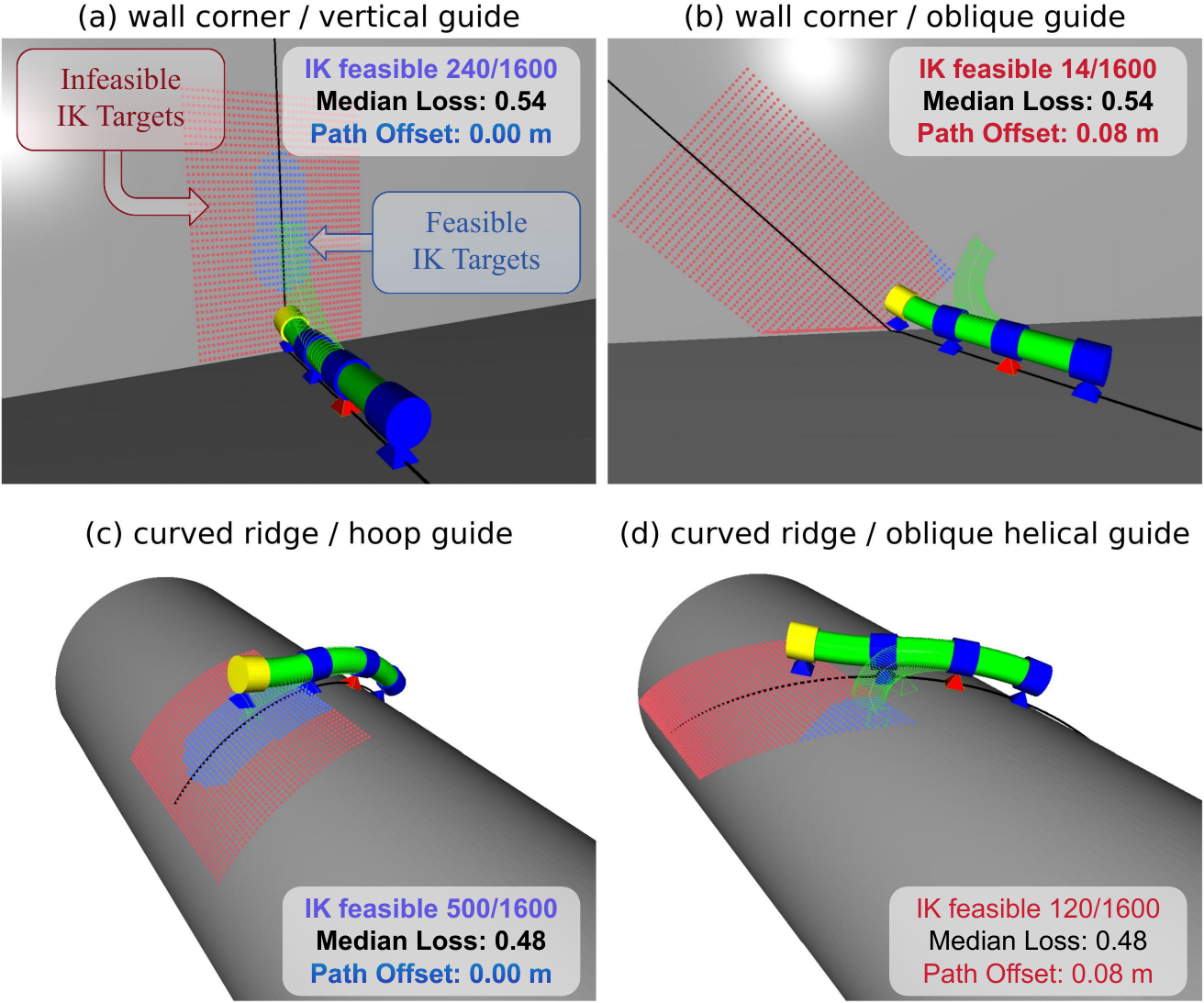}
  \vspace{-0.5mm}
  \caption{Illustration of how the IK-feasible (in blue) and infeasible (in red)
  regions change drastically under different guiding paths (black line).}
  \label{fig:guide_feasibility}
  \vspace{-1mm}
\end{figure}

(IV) \textbf{Effect of guide-path geometry}: Guide geometry changes the
downstream IK-feasible region. At the wall corner, changing the guide from
a vertical path to an oblique path reduces the IK-feasible count from
$240/1600$ to $14/1600$. On the curved ridge, changing the guide from a
hoop-like path to an oblique helical path reduces the count from
$500/1600$ to $120/1600$. In both pairs, the guide with the large path
offset at $0.08\,\mathrm{m}$ yields a \emph{nearly empty}
feasible target set. As shown in Fig.~\ref{fig:guide_feasibility}, this is consistent with~\eqref{eq:guide-cost}: keeping the guide closer to
feasible contact configurations improves both IKHS and PaHS.

(V) \textbf{Embedding ablation}: The learned embedding improves hit rate/PT
over random and pooled PCA features: $87.9\%/2.64\,$s versus
$77.5\%/3.25\,$s and $62.1\%/5.17\,$s in seen environments, and
$80.5\%/2.88\,$s versus $64.3\%/4.09\,$s and $58.6\%/4.29\,$s in unseen
environments (Table~\ref{tab:metric-ablation}); final feasibility remains
enforced by IK refinement.
\begin{table}[t!]
\centering
\caption{Comparison with Baselines.}
\label{tab:compare_ablation}
\footnotesize
\renewcommand{\arraystretch}{0.95}
\setlength{\tabcolsep}{1pt}
\begin{tabular*}{\columnwidth}{@{\extracolsep{\fill}}lcccccccccccc@{}}
\toprule
\multirow{2}{*}{Method}
& \multicolumn{3}{c}{Succ. (/20)}
& \multicolumn{3}{c}{PT (s)}
& \multicolumn{3}{c}{ES}
& \multicolumn{3}{c}{DC} \\
\cmidrule(lr){2-4}\cmidrule(lr){5-7}\cmidrule(lr){8-10}\cmidrule(lr){11-13}
& S1 & S2 & S3 & S1 & S2 & S3 & S1 & S2 & S3 & S1 & S2 & S3 \\
\midrule
\textbf{PaHS}   & \textbf{20} & \textbf{20} & \textbf{20}
                & \textbf{1.7} & \textbf{1.0} & \textbf{1.8}
                & \textbf{51} & \textbf{35} & \textbf{45}
                & \textbf{68} & \textbf{42} & \textbf{84} \\
\textbf{IKHS}   & \textbf{20} & \textbf{20} & \textbf{20}
                & 20.4 & 11.9 & 15.6
                & 64 & 47 & 56
                & 71 & 43 & \textbf{80} \\
PaHS-NF         & 10 & 18 & 11
                & 100.6 & 20.8 & 90.6
                & 125 & 51 & 112
                & 130 & 56 & 128 \\
PWTO            & 9 & 12 & 10
                & 129.2 & 98.6 & 117.7
                & 148 & 118 & 137
                & 180 & 139 & 178 \\
RPR             & 12 & 15 & 3
                & 80.6 & 50.6 & 170.1
                & 109 & 76 & 174
                & 126 & 83 & 174 \\
DBS             & 8 & 14 & 3
                & 161.3 & 146.7 & 180.7
                & 92 & 105 & 176
                & 192 & 189 & 184 \\
\bottomrule
\end{tabular*}
\vspace{-2mm}
\end{table}
\begin{figure}[t!]
  \centering
  \includegraphics[width=1.00\linewidth]{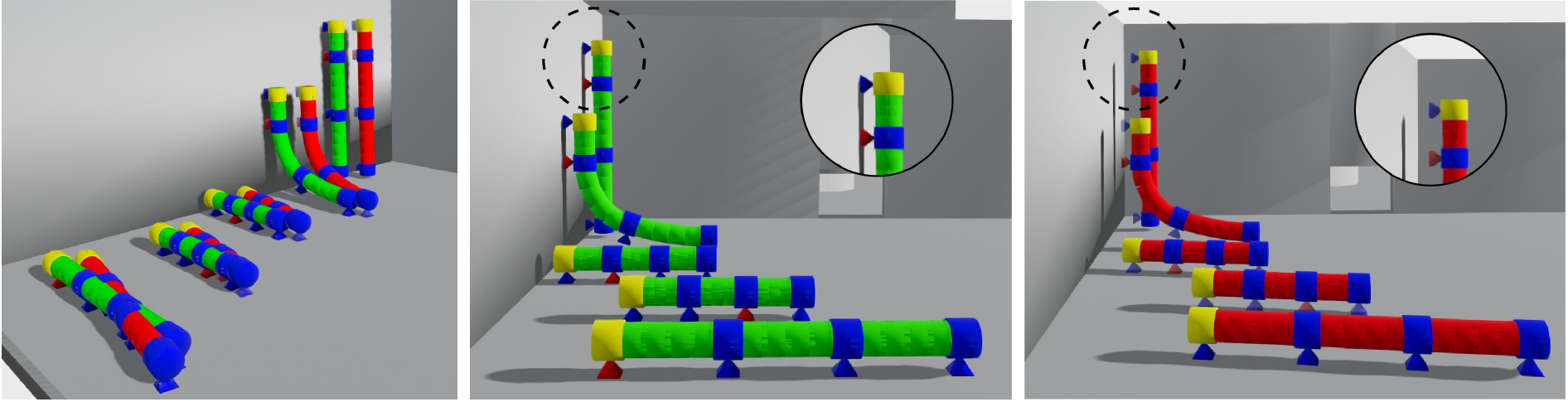}
  \caption{Failure cases for the baseline \textbf{RPR}: without IK refinement,
  small perturbations can turn a feasible wall-transition plan (green) into an
  infeasible one (red) that violates the suction constraint.}
  \label{fig:DIR_failure}
  \vspace{-0.25mm}
\end{figure}

\subsection{Baseline Comparison}\label{subsec:ablation}

PaHS is compared with three controlled baselines and one ablation. Retrieved
primitive replay (\textbf{RPR}) directly replays retrieved motions without online
adaptation or refinement~\cite{phillips2012egraphs}; primitive-warmstarted trajectory
optimization (\textbf{PWTO}) uses them as warm starts for long-horizon optimization
over adhesion and deformation~\cite{bernTrajectoryOpt2019}; discrete beam search
(\textbf{DBS}) searches discretized deformation/adhesion candidates; and \textbf{PaHS-NF}
disables IKHS fallback. Retrieval-based methods use the same top-$50$ library candidates unless noted.

\begin{figure}[t!]
  \centering
  \includegraphics[width=0.93\linewidth]{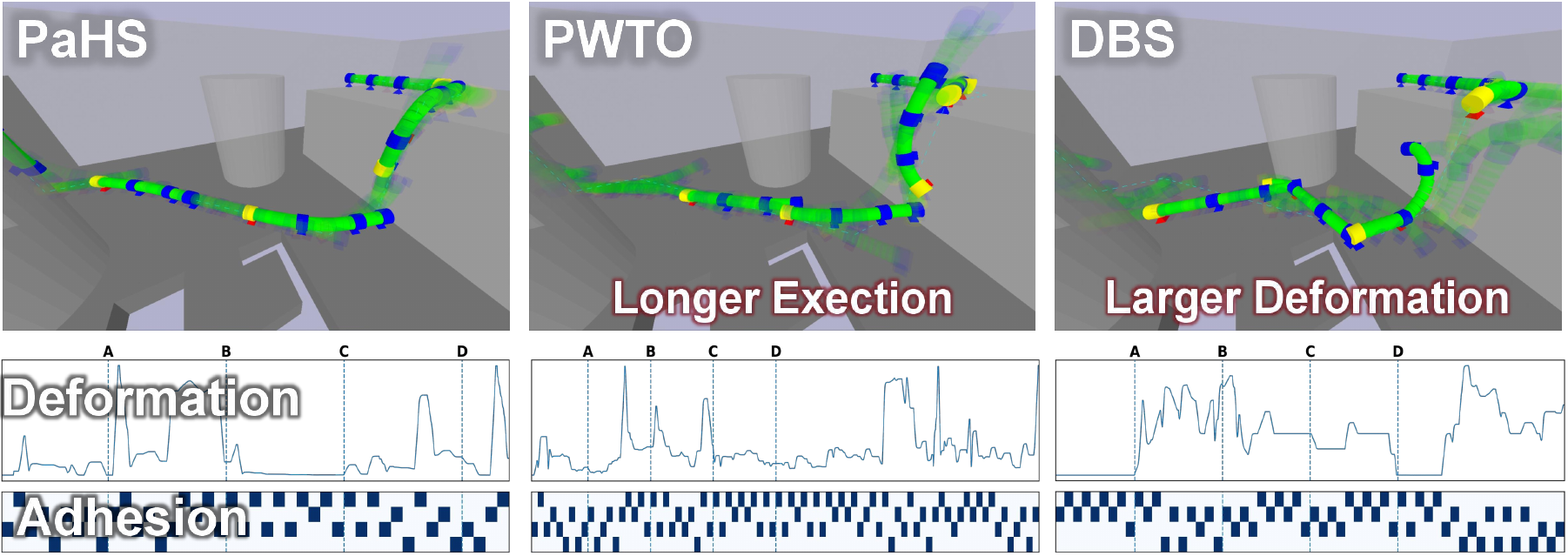}
  \vspace{-1mm}
  \caption{Resulting trajectories under baselines \textbf{PaHS}, \textbf{PWTO},
  and \textbf{DBS}.}
  \label{fig:PWTO_es_compare}
  \vspace{-0.25mm}
\end{figure}

As summarized in Table~\ref{tab:compare_ablation}, PaHS provides the best overall
trade-off: it solves all $60/60$ trials, matching IKHS while reducing mean PT from
$16.0$ to $1.5\,\mathrm{s}$ and achieving the lowest ES in all scenarios, with mean
ES reduced from $55.7$ to $43.7$ without increasing mean DC. PaHS-NF and RPR succeed in only
$39/60$ and $30/60$ trials, showing that retrieval or direct replay alone is
insufficient (Fig.~\ref{fig:DIR_failure}). PWTO and DBS achieve $31/60$ and $25/60$
successes with mean PTs of $115.2$ and $162.9\,\mathrm{s}$. Together with their
larger ES/DC and the trajectories in Fig.~\ref{fig:PWTO_es_compare}, these results
support combining retrieval, IK refinement, and IKHS fallback.

\begin{figure}[t!]
  \centering
  \includegraphics[width=0.96\linewidth]{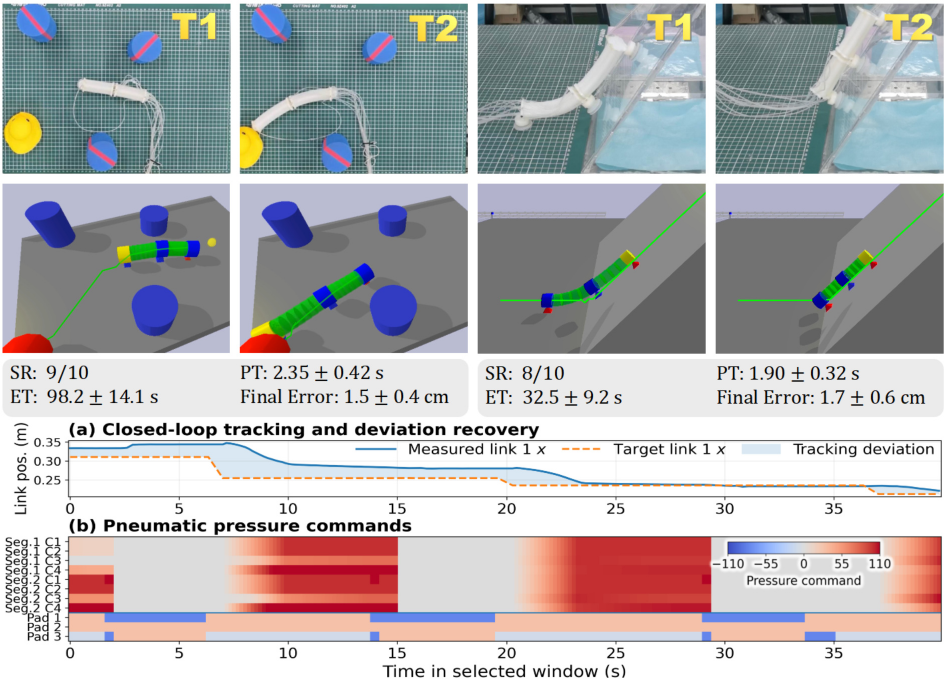}
  \vspace{-0.5mm}
  \caption{
  Snapshots during the planar navigation and the $45^\circ$-slope transition,
  together with the corresponding closed-loop tracking and pressure commands.
  }
  \vspace{-0.25mm}
  \label{fig:hardware}
\end{figure}

\subsection{Hardware Experiments}\label{subsec:hardware}

Hardware experiments use a $60\,\mathrm{cm}\times40\,\mathrm{cm}$ workspace
and a pneumatic robot with two soft segments, three suction pads, and four
chambers per segment. Nine motion-capture markers provide state feedback to
\texttt{PyBullet} and the online planner; clipped integral pressure control uses
$K_i^{\texttt{rot}}=2.0$, $K_i^{\texttt{len}}=5.0$, with saturation at $100$ units.
Two tasks are evaluated (Fig.~\ref{fig:hardware}): randomized planar obstacle
avoidance and a plane-to-slope transition onto a $45^\circ$ incline. The robot
succeeds in $9/10$ planar trials while correcting execution mismatch through online
replanning (mean PT $2.35\,$s; ET $98.2\,$s). In the climbing task, it succeeds in
$8/10$ trials and completes the transition with contact-aware gaits (mean PT $1.9\,$s;
ET $32.5\,$s). Cross-plane motions are more sensitive to imperfect adhesion sealing,
state-estimation error, pneumatic delay, and path-dependent deformation. Moderate
deviations can be recovered through feedback and online replanning, whereas larger
accumulated errors may drive the realized state away from $\mathcal{S}_{\texttt{D}}$,
causing oscillation and preventing recovery.

\section{Conclusion}\label{sec:conclusion}
This paper presented a planning and control framework for multi-segment
suction-based soft robots on complex 3D surfaces. The central idea is an
adhesion-induced block-wise planning abstraction, implemented through IKHS and
further accelerated by primitive proposals in PaHS. Future work will focus on
broader hardware validation and extended application settings.

\bibliographystyle{IEEEtran}
\bibliography{contents/references}

\clearpage
\setlength{\textfloatsep}{6pt plus 1pt minus 1pt}
\setlength{\dbltextfloatsep}{7pt plus 1pt minus 1pt}
\setlength{\floatsep}{5pt plus 1pt minus 1pt}
\setlength{\intextsep}{5pt plus 1pt minus 1pt}
\setlength{\abovecaptionskip}{2pt}
\setlength{\belowcaptionskip}{-2pt}
\setlength{\abovedisplayskip}{4pt}
\setlength{\belowdisplayskip}{4pt}
\setlength{\abovedisplayshortskip}{3pt}
\setlength{\belowdisplayshortskip}{3pt}
\twocolumn[
\begin{center}
{\Large\bfseries Supplementary Material for}\par
\vspace{0.35em}
{\large\bfseries FlexWorm: Primitive-augmented Hybrid Contact-motion Planning\par
for Suction-based Multi-segment Deformable Robots}\par
\vspace{0.6em}
Zili Tang, Tiecheng Guo, Qinyue Zhang, and Meng Guo
\end{center}
\vspace{0.5em}
]
\setcounter{figure}{0}
\setcounter{table}{0}
\setcounter{equation}{0}
\renewcommand{\thefigure}{S\arabic{figure}}
\renewcommand{\thetable}{S\arabic{table}}
\renewcommand{\theequation}{S\arabic{equation}}
\renewcommand{\theHfigure}{S\arabic{figure}}
\renewcommand{\theHtable}{S\arabic{table}}
\renewcommand{\theHequation}{S\arabic{equation}}

\newcommand{\SD}{\mathcal{S}_{\texttt{D}}}
\newcommand{\pospart}[1]{\left[#1\right]_{+}}

\setcounter{subsection}{0}
\setcounter{subsubsection}{0}

The following supplementary material expands the implementation details summarized in the main paper and provides additional validation. Sections~A--D specify the guiding-path cost, block-wise IK objective, load--deformation feasibility model, and primitive-retrieval training pipeline, respectively; Sec.~E provides additional deformation and hardware analyses. Shared numerical parameters follow Table~\ref{tab:impl-params} and are not repeated unless needed for clarity.

\subsection{Guiding Surface Path Cost}
\label{subsec:supp-guide}

\subsubsection{Augmented Near-surface Graph}

Section~III-A1 summarizes the guide construction and its role in IKHS. This section gives the exact implementation of the augmented near-surface graph and the individual penalty terms in the edge cost. The guide is computed on $\bar{\mathcal{G}}_{\texttt{s}}$, whose locations $\hat{\mathbf{p}}$ are lattice points in the near-surface band, i.e.,
\begin{equation}
\mathcal{V}_s
=
\{\hat{\mathbf{p}}: d_{\min}\le D(\hat{\mathbf{p}})\le d_{\max}\},
\label{eq:supp-near-surface-band}
\end{equation}
where $D(\cdot)$ is the signed-distance field,
$d_{\min}=0.005\,\mathrm{m}$, and
$d_{\max}=0.025\,\mathrm{m}$. Because edge costs depend on the local approach
direction, the search state is augmented as
$\overline v=(\hat{\mathbf{p}},\delta)$, where $\delta$ is a quantized tangent
direction. A short ancestor history is also retained to evaluate recent
turning and surface-transition geometry.
For an edge from step~$t$ to step~$t+1$, the guide cost is given by:
\begin{equation}
\begin{split}
c_{t\rightarrow t+1}
={}&w_\ell d_{t\rightarrow t+1}
+w_{\texttt{lift}}\phi_{\texttt{lift}}
+w_{\texttt{sharp}}\phi_{\texttt{sharp}}\\
&+w_{\texttt{n}}\phi_{\texttt{normal}}
+w_{\texttt{b}}\phi_{\texttt{bend}}
+w_{\texttt{nc}}\phi_{\texttt{nc}},
\end{split}
\label{eq:supp-guide-cost}
\end{equation}
where the weights are given in Table~\ref{tab:impl-params}. Overall scales are
absorbed into these external weights, and the internal coefficients
are normalized by their largest value.

\subsubsection{Length, Lift, and Surface-normal Change}

The geometric term is the metric length of the 26-neighbor step,
$d_{t\rightarrow t+1}
=\|\hat{\mathbf{p}}_{t+1}-\hat{\mathbf{p}}_t\|_2$. The lift term is
$\phi_{\texttt{lift}}(\hat{\mathbf{p}}_{t+1})
=\pospart{D(\hat{\mathbf{p}}_{t+1})-d_{\min}}$, which discourages unnecessary
lift within the near-surface band. Let
$\mathbf{n}_t=\mathbf{n}(\hat{\mathbf{p}}_t)$ and
$\Delta\theta_n=\cos^{-1}(
\operatorname{clip}(\mathbf{n}_t^\top\mathbf{n}_{t+1},-1,1))$. The
normal-change penalty is
$\phi_{\texttt{normal}}=\Delta\theta_n+(\Delta\theta_n)^2$, which discourages
rapid surface-normal variation and repeated floor--wall--ceiling transitions.

\subsubsection{Local Sharpness}

Let $\overline{\mathbf{t}}_{t+1}$ be the unit-normalized exponential moving average
of guide steps with update coefficient $0.2$, and let
$\mathbf{s}_{t+1}$ be the transverse direction obtained by normalizing
$\mathbf{n}_{t+1}\times\overline{\mathbf{t}}_{t+1}$. Degenerate cases with a
near-zero denominator are skipped in this local sharpness test. The SDF is
probed on both transverse sides of the guide at five voxel cells,
i.e., $5\Delta=0.05\,\mathrm{m}$ by:
\begin{equation}
\begin{aligned}
\phi_{\texttt{sharp}}
=
\sum_{\sigma\in\{-1,+1\}}
\mathds{1}\!\Big(
&\big|D(\hat{\mathbf{p}}_{t+1}+5\Delta\sigma\mathbf{s}_{t+1})\\
&-D(\hat{\mathbf{p}}_{t+1})\big|>\tau_D
\Big),
\end{aligned}
\label{eq:supp-sharp-cost}
\end{equation}
where $\Delta$ is the voxel resolution and $\tau_D=d_{\max}/2$. This term
penalizes abrupt transverse SDF variation around the guide.

\subsubsection{Accumulated In-plane Bending}

The previous and current step directions are projected into the tangent plane
at $\hat{\mathbf{p}}_{t+1}$. Let the angle between the projected directions be
$\Delta\theta_t$. Small direction changes below $\theta_0=\pi/20$ are ignored.
With discount factor $\beta_a=0.9$, the accumulated turning history is updated
as $a_{t+1}=\beta_a a_t+\pospart{\Delta\theta_t-\theta_0}$ with $a_0=0$. The
bending penalty is
$\phi_{\texttt{bend}}=0.005a_{t+1}
+\pospart{a_{t+1}-\pi/4}^{2}$, which prevents a sequence of individually small
turns from hiding a large accumulated bend.

\subsubsection{Non-coplanar Transition Penalty}

For a recent ancestor~$k$, let
$\mathbf{n}_k=\mathbf{n}(\hat{\mathbf{p}}_k)$,
$\mathbf{d}_k=(\hat{\mathbf{p}}_{t+1}-\hat{\mathbf{p}}_k)/
\|\hat{\mathbf{p}}_{t+1}-\hat{\mathbf{p}}_k\|_2$,
and let~$\mathbf{c}_k$ be the normalized vector
$\mathbf{d}_k\times\mathbf{n}_k$. Ancestors that lead to near-zero denominators
are skipped. If the new surface normal is coplanar with the travel direction
and the previous normal, then $|\mathbf{c}_k^\top\mathbf{n}_{t+1}|$ is small.
The normalized history penalty is given by:
\begin{equation}
\phi_{\texttt{nc}}
=
\frac{1}{K_{\texttt{nc}}}
\sum_{k\in\mathcal{H}_{K_{\texttt{nc}}}}
\mathds{1}(|\mathbf{c}_k^\top\mathbf{n}_{t+1}|>0.1)
|\mathbf{c}_k^\top\mathbf{n}_{t+1}|,
\label{eq:supp-noncoplanar}
\end{equation}
where $\mathcal{H}_{K_{\texttt{nc}}}$ contains up to
$K_{\texttt{nc}}=15$ recent ancestors and is evaluated only near a surface
switch. With the external weight $w_{\texttt{nc}}=30$, this term penalizes
recent non-coplanar surface transitions in the guide.


\subsection{Block-wise IK Target-pose Objective}
\label{subsec:supp-target-objective}

\subsubsection{Target Error Terms}

The compact target-pose objective in Eq.~\eqref{eq:perblock-ik} of the main paper is expanded here into its position and orientation residuals. For a sampled suction target $\mathbf{x}^\star=(\mathbf{p}^\star,\mathbf{q}^\star)$, the desired suction normal is defined as:
\begin{equation}
\mathbf{n}^\star
=
R(\mathbf{q}^\star)\mathbf{e}_{\texttt{x}},
\label{eq:supp-target-normal}
\end{equation}
where $R(\mathbf{q})$ is the rotation matrix associated with quaternion
$\mathbf{q}$, and $\mathbf{e}_{\texttt{x}}=(1,0,0)^\top$ is the local suction-axis unit
vector. For the reached target-link state
$\mathbf{x}=(\mathbf{p},\mathbf{q})$,
the target quaternion is sign-aligned as:
\begin{equation}
\widetilde{\mathbf{q}}^\star
=
\begin{cases}
\mathbf{q}^\star, & \mathbf{q}^{\top}\mathbf{q}^\star\ge 0,\\
-\mathbf{q}^\star, & \mathbf{q}^{\top}\mathbf{q}^\star<0,
\end{cases}
\label{eq:supp-q-align}
\end{equation}
where the sign alignment removes the double-cover ambiguity of unit
quaternions before evaluating the residual. The normal-distance,
tangential-distance, normal-orientation,
and residual-orientation errors are computed as follows:
\begin{equation}
\begin{aligned}
e_{\texttt{n}}
&=
(\mathbf{p}-\mathbf{p}^\star)^\top\mathbf{n}^\star,\\
\mathbf{e}_{\texttt{t}}
&=
(\mathbf{p}-\mathbf{p}^\star)-e_{\texttt{n}}\mathbf{n}^\star,\\
\mathbf{e}_{\texttt{no}}
&=
R(\mathbf{q})\mathbf{e}_{\texttt{x}}-\mathbf{n}^\star,\\
\mathbf{e}_{\texttt{o}}
&=
\mathbf{q}-\widetilde{\mathbf{q}}^\star,
\end{aligned}
\label{eq:supp-target-errors}
\end{equation}
where $e_{\texttt{n}}$ and $\mathbf{e}_{\texttt{t}}$ decompose the position
error into normal and tangential components, $\mathbf{e}_{\texttt{no}}$
measures suction-axis misalignment, and $\mathbf{e}_{\texttt{o}}$ penalizes the
remaining quaternion mismatch.

\subsubsection{Weighted Target-pose Loss}

The target-pose loss used in block-wise IK is defined as:
\begin{equation}
\mathcal{L}_{\mathrm{tar}}
=
w_{\mathrm{nd}}e_{\mathrm{n}}^2
+w_{\mathrm{td}}\|\mathbf{e}_{\mathrm{t}}\|_2^2
+w_{\mathrm{no}}\|\mathbf{e}_{\mathrm{no}}\|_2^2
+w_{\mathrm{o}}\|\mathbf{e}_{\mathrm{o}}\|_2^2,
\label{eq:supp-targetloss}
\end{equation}
where the numerical weights are given in Table~\ref{tab:impl-params}. The
larger normal-distance and normal-orientation weights prioritize seal
formation. Tangential position and residual rotation are assigned smaller
weights because a short free block cannot generally satisfy an arbitrary
six-DOF target exactly.


\subsection{Load--deformation Feasible Set}
\label{subsec:supp-sd}

\subsubsection{Offline Gravity-conditioned Construction}

Section~III-A3 summarizes the gravity-conditioned feasible family $\SD(\gamma)$. This section gives its calibrated construction, the load computation used for full-chain validation, and the convex slice used inside block-wise IK. Approximately $10^3$ hardware measurements calibrate a finite-element segment model, which generates about $5\times10^5$ quasi-static samples over segment length, the two-dimensional bending-curvature vector $\boldsymbol{\kappa}\in\mathbb{R}^2$, distal load moment, and proximal-frame gravity angle. The gravity angle $\gamma$ is discretized into $15^\circ$ intervals. For each interval, approximately $6\times10^4$ uncertain near-boundary samples are discarded before fitting a conservative inner convex polytope in $(\overline M,L,\boldsymbol{\kappa})$ space. Thus, for segment~$i$ at hybrid step~$h$, the planning constraint is given by:
\begin{equation}
(\overline M_i^h,L_i^h,\boldsymbol{\kappa}_i^h)
\in \SD(\gamma_i^h),
\label{eq:supp-sd-membership}
\end{equation}
where $L_i^h$ and $\boldsymbol{\kappa}_i^h$ are the segment deformation
parameters, $\overline M_i^h$ is the equivalent distal load moment, and
$\gamma_i^h$ is the gravity angle in the proximal fixed-end frame.

\subsubsection{Full-chain Load Validation}

The equivalent load moment is evaluated from the current full-chain
configuration. For an attached segment anchor~$a$, the dominant
gravity-induced moment is approximated by:
\begin{equation}
\mathbf{M}_a
\triangleq
\sum_{j\in\mathcal{D}(a)}
(\mathbf{p}_j-\mathbf{p}_a)\times m_j\mathbf{g},
\qquad
\overline M_a\triangleq\|\mathbf{M}_a\|_2,
\label{eq:supp-load-moment}
\end{equation}
where $\mathcal{D}(a)$ denotes the distal links supported by anchor~$a$,
$m_j$ is the lumped mass of link~$j$, and $\mathbf{g}$ is gravitational
acceleration. The same feasible family is used for all nominally identical
segments. Multi-segment coupling is still checked at the full-chain level:
after each candidate transition, the complete chain state is recomputed, the
relevant load moments and gravity angles are updated, and all affected or
load-supporting segments are checked against~\eqref{eq:supp-sd-membership},
together with closure, suction-pose, and collision constraints.

\subsubsection{Frozen Convex Section for Block-wise IK}

During block-wise numerical IK, directly updating
$\overline M_i$ and $\gamma_i$ inside every nonlinear iteration would make the
inner problem less stable. Therefore, the implementation uses a
load-conditioned convex section frozen at the parent state. For an affected
segment~$i$ in a free block, the parent load moment and gravity angle
yield the deformation slice below:
\begin{equation}
\mathcal{C}_i^0
\triangleq
\left\{
(L,\boldsymbol{\kappa})
\;\middle|\;
(\overline M_i^0,L,\boldsymbol{\kappa})
\in \SD(\gamma_i^0)
\right\},
\label{eq:supp-frozen-slice}
\end{equation}
where $\mathcal{C}_i^0$ is the feasible deformation slice used only inside
the local IK optimization. Because $\SD(\gamma_i^0)$ is represented by a
convex inner polytope, $\mathcal{C}_i^0$ is convex and can be imposed
efficiently in the block IK solve. Once the block solutions are fused into a
full child state, the child is retained only if the recomputed quantities
satisfy the following condition:
\begin{equation}
(\overline M_i',L_i',\boldsymbol{\kappa}_i')
\in \SD(\gamma_i'),
\qquad
\forall i\in\mathcal{I}_{\texttt{aff}},
\label{eq:supp-final-sd-check}
\end{equation}
where $\mathcal{I}_{\texttt{aff}}$ includes the optimized segments and any
supporting segments whose distal load changes.

\begin{figure}[t!]
\centering
\includegraphics[width=1\linewidth]{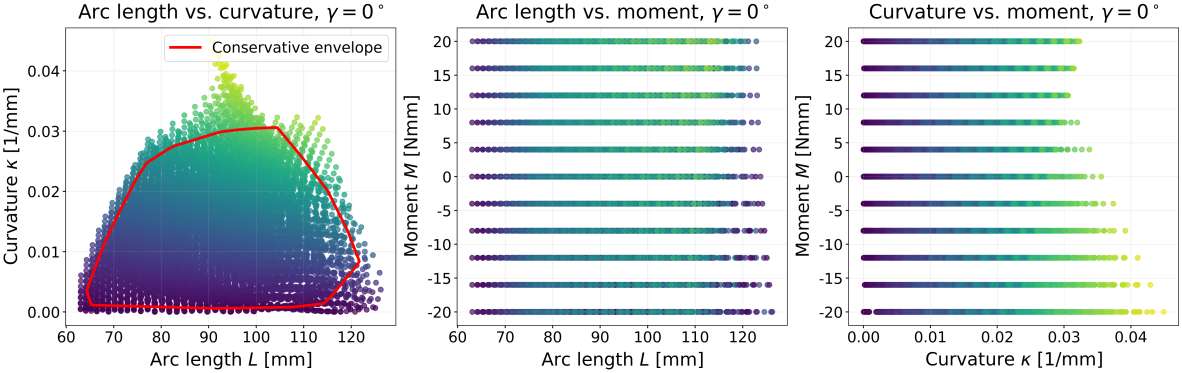}\\[1mm]
\includegraphics[width=1\linewidth]{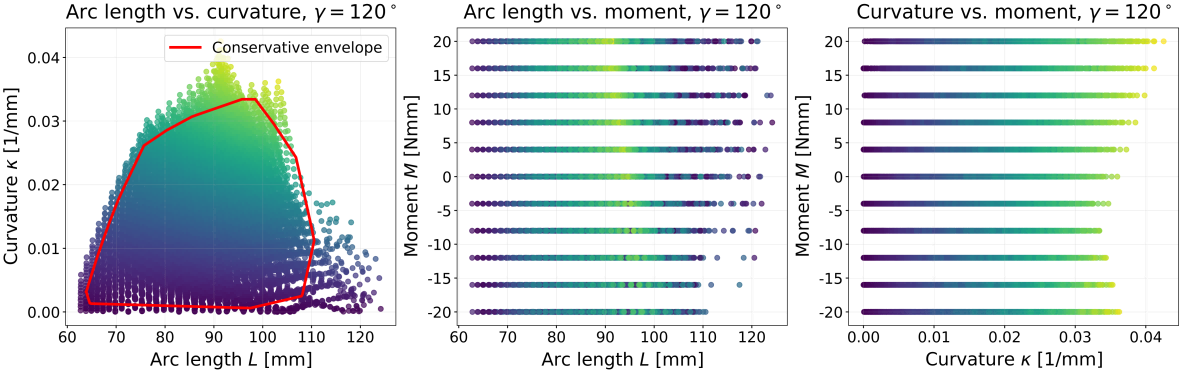}
\caption{Projected quasi-static samples used to construct $\SD(\gamma)$ for $\gamma=0^\circ$ and $\gamma=120^\circ$. For visualization, the two-dimensional curvature vector is projected to its magnitude $\kappa\triangleq\|\boldsymbol{\kappa}\|_2$; the planner retains the full vector $\boldsymbol{\kappa}\in\mathbb{R}^2$. The panels show the $(L,\kappa)$, $(L,\overline M)$, and $(\kappa,\overline M)$ projections. The red curves indicate projected conservative bounds where shown.}
\label{fig:supp-sd-proj}
\end{figure}

\begin{figure}[t!]
\centering
\includegraphics[width=0.48\linewidth]{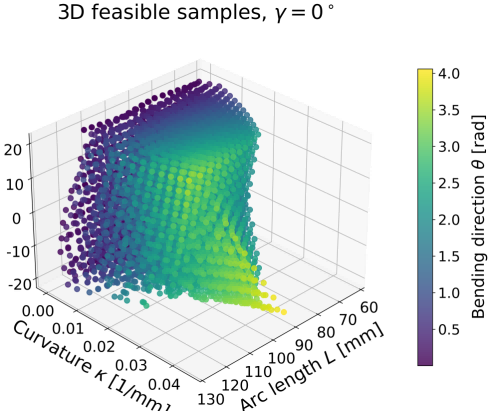}
\includegraphics[width=0.48\linewidth]{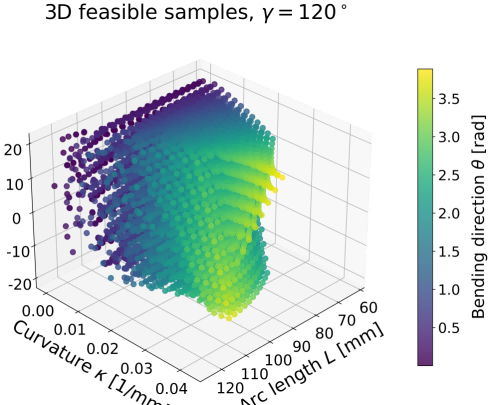}
\caption{Three-dimensional magnitude projections of the fitted gravity-conditioned feasible sets for $\gamma=0^\circ$ and $\gamma=120^\circ$, using $\kappa=\|\boldsymbol{\kappa}\|_2$ for visualization. The planner-side feasible set itself is represented in $(\overline M,L,\boldsymbol{\kappa})$ space.}
\label{fig:supp-sd-3d}
\end{figure}

\subsubsection{Conservativeness and Interpretation}

The fitted feasible sets for two representative gravity angles are visualized
in Fig.~\ref{fig:supp-sd-proj}. The fitted polytopes are conservative inner
approximations, so they may reject some quasi-statically feasible states near
the boundary. Conversely, membership in $\SD(\gamma)$ should be interpreted as
calibrated quasi-static feasibility, not guaranteed closed-loop trackability
under pneumatic delay, sealing uncertainty, hysteresis, or state-estimation
error.
The same fitted sets are also shown through three-dimensional magnitude projections in Fig.~\ref{fig:supp-sd-3d}. These views make the conservativeness of the inner approximation clearer, while the planner retains the full two-dimensional curvature vector rather than the scalar magnitude used for visualization.

\begin{figure*}[t!]
\centering
\includegraphics[width=0.94\textwidth]{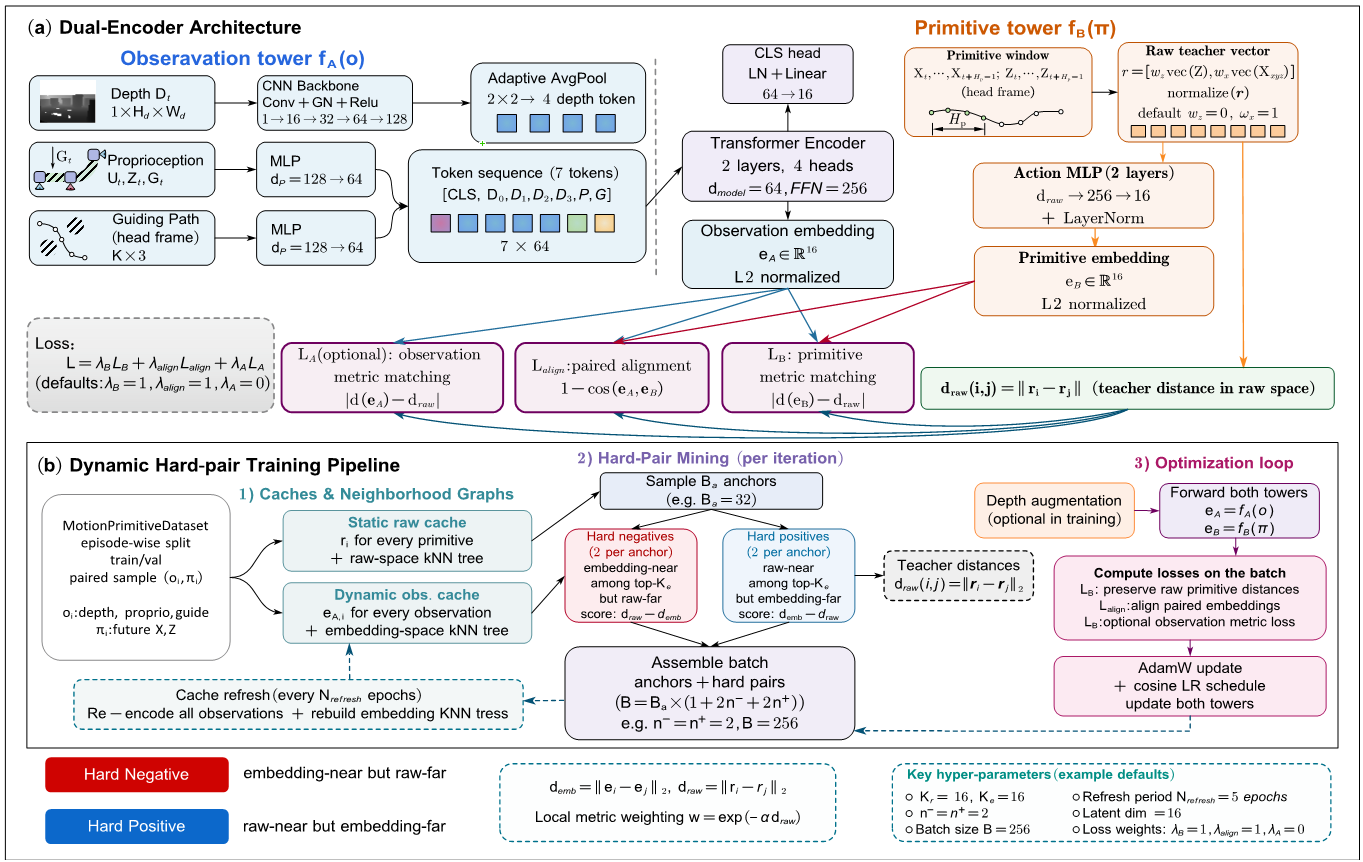}
\vspace{-1mm}
\caption{PaHS primitive-retrieval model and training protocol. The observation
tower embeds the local depth, proprioception, gravity, and guide-window inputs,
while the action tower embeds validated IKHS primitive continuations. Retrieved
primitives are replayed, refined, and accepted after
validation.}
\label{fig:supp-pahs-arch}
\end{figure*}


\subsection{Primitive Retrieval Model and Training Protocol}
\label{subsec:supp-pahs}

\subsubsection{Primitive Records}

Section~III-B describes PaHS as a retrieval-augmented proposal mechanism inside the same IKHS search and validation loop. The learned module only accelerates proposal generation: every retrieved continuation is still refined and checked for suction pose, collision, IK loss, and load--deformation feasibility, with standard IKHS fallback retained. This section specifies the exact primitive records, encoder inputs, metric target, hard-pair mining, and episode-level training split used in the reported implementation.

The primitive library is generated from successful IKHS rollouts. A sample at
hybrid step~$h$ pairs the local observation
\begin{equation}
o_h=(I_h^{\texttt{depth}},\mathbf{U}^h,\mathbf{Z}^h,
\mathbf{g}^h,\mathcal{G}^h),
\label{eq:supp-pahs-obs}
\end{equation}
with the subsequent validated primitive of horizon $H_{\texttt{p}}=4$,
\begin{equation}
\pi_h=
(\mathbf{X}^{h:h+H_{\texttt{p}}},
\mathbf{U}^{h:h+H_{\texttt{p}}-1},
\mathbf{Z}^{h:h+H_{\texttt{p}}}),
\label{eq:supp-pahs-primitive}
\end{equation}
where $I_h^{\texttt{depth}}$ is a local depth image,
$\mathbf{U}^h$ and $\mathbf{Z}^h$ are the current deformation and adhesion-mode
variables, $\mathbf{g}^h$ contains the link-wise gravity directions, and
$\mathcal{G}^h$ is a short guide look-ahead window. Guide points and primitive
poses are represented in the current head frame. Each primitive record also
stores the corresponding segment twists, adhesion modes, and IK targets, so
that the retrieved primitive can be replayed and refined rather than directly
accepted.

\subsubsection{Dual-tower Embedding Model}

Retrieval uses the dual-tower model summarized in Sec.~III-B2 and Fig.~6 of the main paper. For completeness, the exact encoder inputs are specified here. The observation tower $E_{\texttt{obs}}$ maps $o_h$ to a normalized query embedding $\mathbf{e}_{\texttt{obs}}\in\mathbb{R}^{16}$ and combines a $56{\times}56$ depth-image backbone, MLP encoders for proprioception and gravity, a guide-window encoder with $K_g=50$ guide points, and a two-layer four-head Transformer fusion module. The action tower $E_{\texttt{act}}$ maps each validated continuation to a normalized primitive key $\mathbf{e}_{\texttt{act}}\in\mathbb{R}^{16}$ through a two-layer MLP applied to the normalized link-position descriptor below:
\begin{equation}
\mathbf{r}(\pi_h)
=
\textsf{norm}\!\left(
\textsf{vec}(\mathbf{X}^{h:h+H_{\texttt{p}}}_{xyz})
\right),
\label{eq:supp-pahs-desc}
\end{equation}
where only link-position geometry is used to define the metric target.
Adhesion modes, segment twists, and IK targets remain in the primitive payload
and are revalidated during online use.
The resulting dual-tower retrieval model and training pipeline are summarized
in Fig.~\ref{fig:supp-pahs-arch}. The figure also clarifies that the learned
embedding is used only for primitive retrieval, while feasibility is still
enforced by the same IK refinement and validation checks as IKHS.

\subsubsection{Training Objective and Data Split}

The training objective aligns observations with their paired primitives while
preserving geometric similarity among primitive trajectories:
\begin{equation}
\mathcal{L}
=
\lambda_{\texttt{act}}\mathcal{L}_{\texttt{act}}^{\mathrm{metric}}
+\lambda_{\texttt{align}}\mathcal{L}_{\texttt{align}}
+\lambda_{\texttt{obs}}\mathcal{L}_{\texttt{obs}}^{\mathrm{metric}},
\label{eq:supp-pahs-loss}
\end{equation}
where
$(\lambda_{\texttt{act}},\lambda_{\texttt{align}},
\lambda_{\texttt{obs}})=(1,1,0)$ in the reported model. The action-tower
metric loss matches primitive-embedding distances to the teacher distances
induced by~\eqref{eq:supp-pahs-desc}, while the alignment loss pulls each
observation query toward the key of its paired primitive. Dynamic hard-pair
sampling is used during training: hard positives are geometrically similar
primitive pairs that remain far in the learned space, and hard negatives are
geometrically different pairs that appear too close in the learned space. The
model is trained with batch size $256$, learning rate $10^{-3}$, weight decay
$10^{-5}$, and $200$ epochs.

The data split is performed at the episode level before fixed-horizon primitive
windows are extracted. Specifically, $500$ successful IKHS episodes are
collected in each of three training scenarios, giving $1500$ episodes and more
than $4\times10^4$ primitive windows. Episodes are split $90/10$ into training
and validation sets before window extraction, preventing overlapping windows
from the same rollout from appearing in both sets. After embedding, indexing,
and duplicate removal, the deployed primitive library contains about $6000$
entries. The final planning evaluations use independent random seeds and are
not drawn from the training or validation episodes.

\subsection{Additional Evidence for Deformation Regularization and Hardware Execution}
\label{subsec:supp-additional}

\begin{figure}[t!]
\centering
\includegraphics[width=1\linewidth]{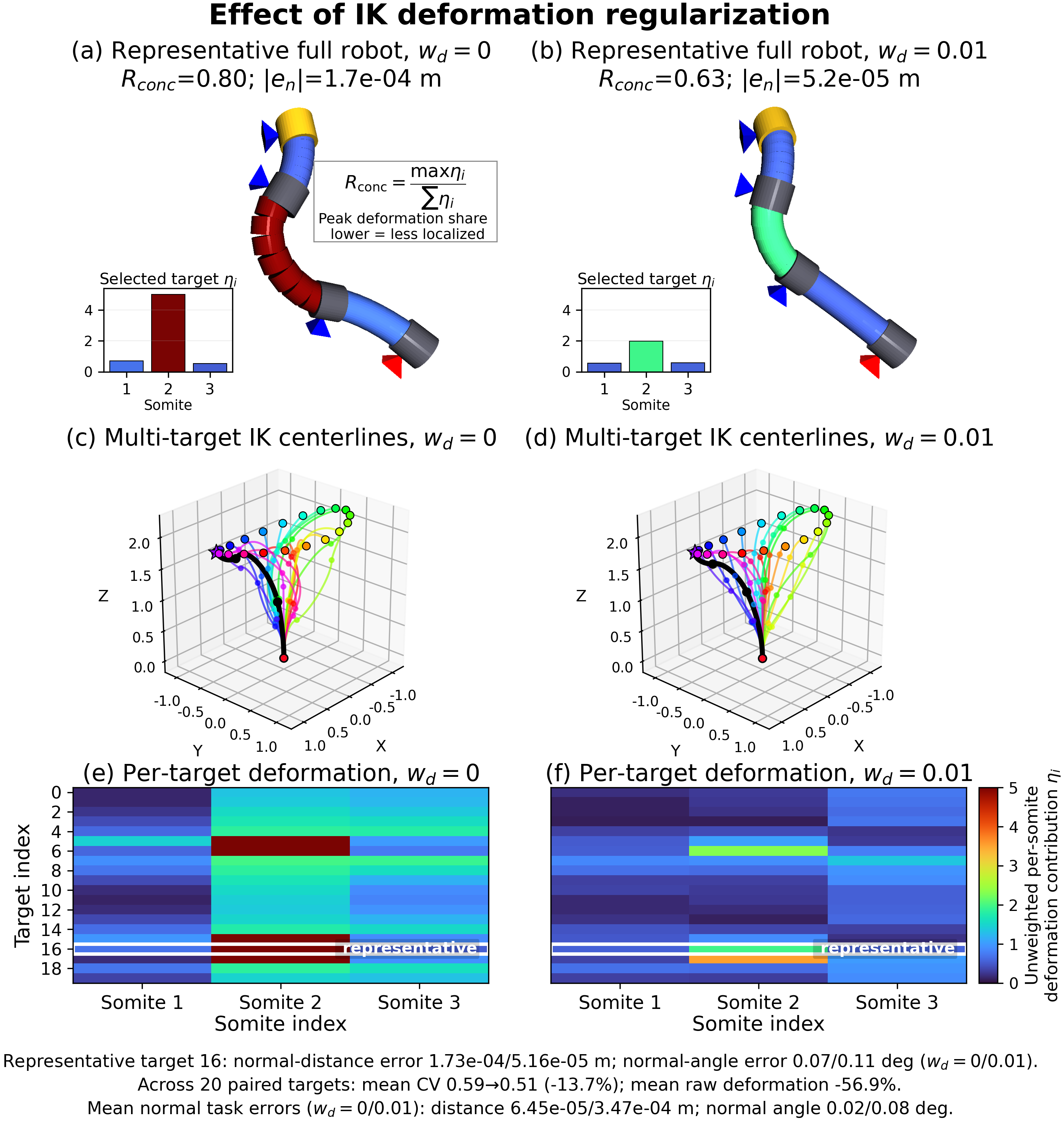}
\caption{Effect of deformation regularization in numerical IK. The upper
panels show representative full-robot solutions and per-segment deformation
contributions $\eta_i$; the middle panels show paired multi-target solution
families; the lower panels show per-target deformation contributions.}
\label{fig:supp-ikdef}
\end{figure}

\subsubsection{IK-level Deformation Regularization}

The deformation cost is used only as a soft preference among feasible solutions; it does not replace the hard suction-pose, collision, closure, or load--deformation constraints. The same normalized extension and bending quantities enter at two levels: $\mathcal{L}_{\texttt{def}}$ in Eq.~\eqref{eq:perblock-ik} biases each local IK solve, while $\mathcal{J}_{\texttt{def}}$ in Eq.~\eqref{eq:ikhs-select} accumulates deformation along the search trajectory. For segment~$i$, let $L_i$ be the segment length, $L_{\min}$ and $L_{\max}$ its length limits, and $\boldsymbol{\theta}_i\triangleq L_i\boldsymbol{\kappa}_i$ the bending-angle vector induced by the curvature $\boldsymbol{\kappa}_i$. The normalized extension and per-segment deformation measure are given by:
\begin{equation}
\rho_i
=
\frac{L_i-L_{\min}}{L_{\max}-L_{\min}},
\qquad
\eta_i=(1+\rho_i)\|\boldsymbol{\theta}_i\|_2^2,
\label{eq:supp-def-measure}
\end{equation}
where $\rho_i$ measures normalized extension and $\eta_i$ summarizes the local bending contribution used by the deformation regularizer. In block-wise IK, $\mathcal{L}_{\texttt{def}}$ uses these quantities to select lower-deformation solutions among contact-compatible target candidates. The search-level $\mathcal{J}_{\texttt{def}}$ accumulates the corresponding normalized extension and squared-bending penalties across transitions, so the planner prefers lower-strain routes without relaxing any feasibility constraint.

A paired numerical-IK ablation over $20$ target poses is shown in
Fig.~\ref{fig:supp-ikdef}. The peak deformation share is defined as:
\begin{equation}
R_{\texttt{conc}}
=
\frac{\textbf{max}_i\{\eta_i\}}{\sum_i\eta_i},
\label{eq:supp-conc-ratio}
\end{equation}
where $R_{\texttt{conc}}$ measures whether the deformation is concentrated in
one segment. In the representative case, enabling the regularizer decreases
$R_{\texttt{conc}}$ from $0.80$ to $0.63$. Across the paired targets, the mean
coefficient of variation decreases from $0.59$ to $0.51$, and the mean raw
deformation decreases by $56.9\%$, while the normal-distance and surface-normal
orientation errors remain negligible. This indicates that the regularizer uses
IK redundancy to distribute deformation, rather than relaxing the suction
target.

\begin{figure}[t!]
\centering
\includegraphics[width=1\linewidth]{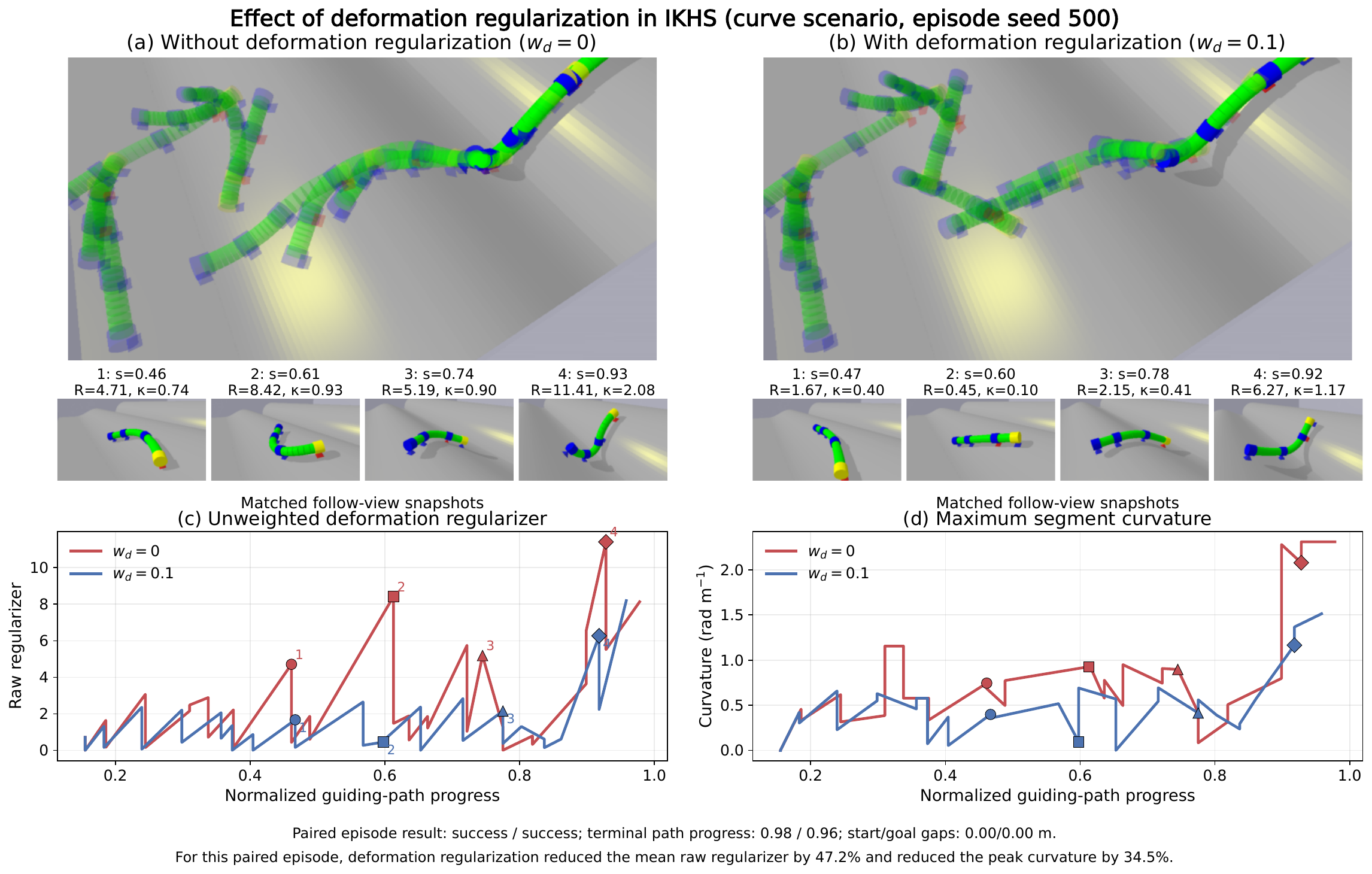}
\caption{Planning-level deformation ablation in the curve scenario. Global
ghost trajectories and matched follow-view frames are shown above the raw
deformation and maximum-curvature traces.}
\label{fig:supp-plandef}
\end{figure}

\subsubsection{Planning-level Deformation Regularization}

The curve-scenario planning runs with the search-level deformation preference disabled and enabled are compared in Fig.~\ref{fig:supp-plandef}. To distinguish this ablation from the IK weight $w_d=10^{-2}$ in Eq.~(14), define the base search weight $\overline w_{\texttt{d}}\triangleq0.1$, so that the adaptive score weight in Eq.~\eqref{eq:ikhs-select} is $w_{\texttt{d}}^h=\alpha_{\texttt{p}}\overline w_{\texttt{d}}$. The figure labels this base search-level ablation parameter as $w_d$ for compactness. Both runs reach nearly identical terminal progress, but the regularized trajectory avoids highly curved states. Enabling $\overline w_{\texttt{d}}=0.1$ decreases the mean raw deformation by $47.2\%$ and the peak segment curvature by $34.5\%$. Thus, the search-level deformation preference improves trajectory quality within the same feasible-transition set rather than changing the task definition or removing hard constraints.

\begin{figure}[t!]
\centering
\includegraphics[width=0.98\linewidth]{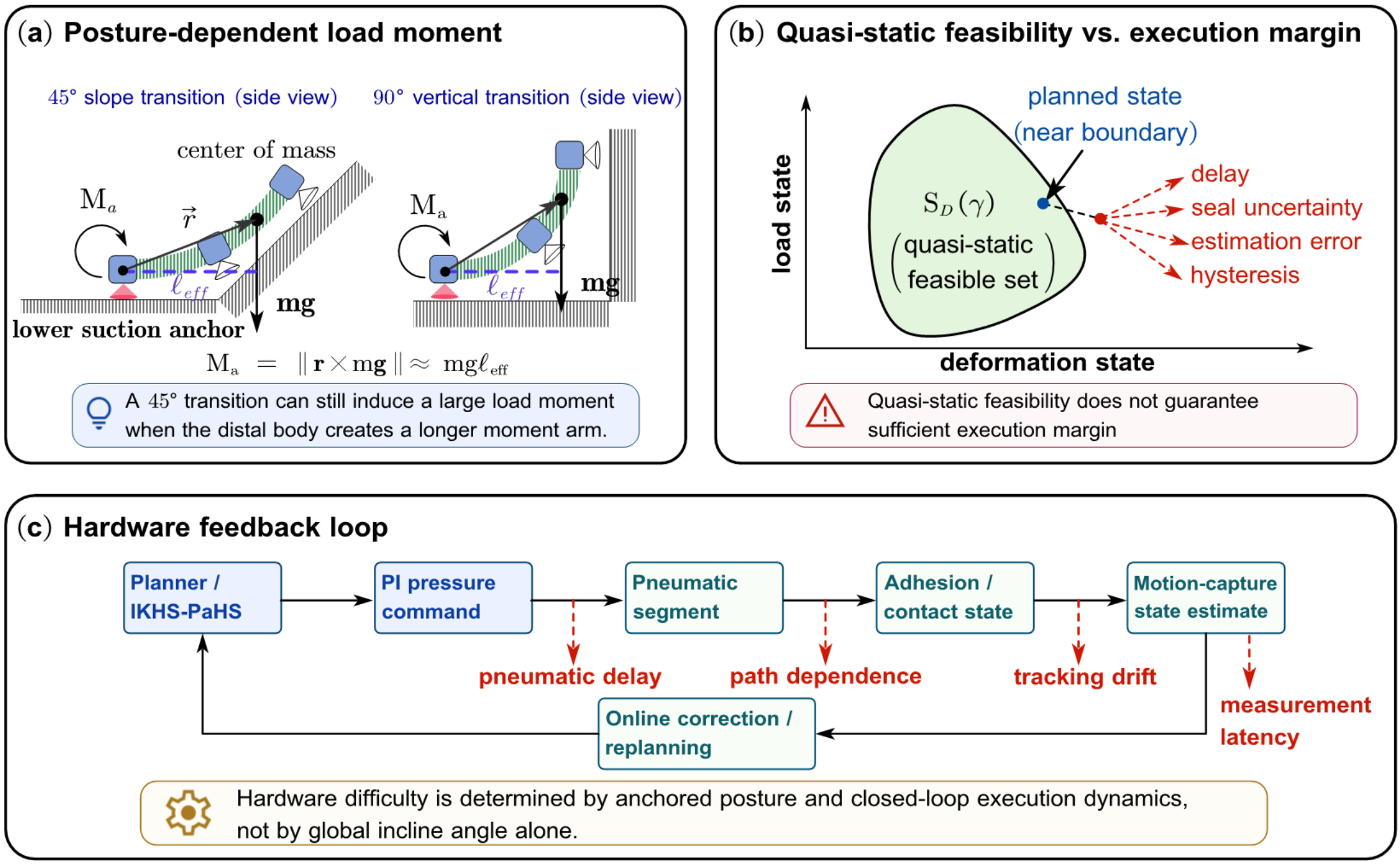}
\caption{Illustration of the hardware execution difficulty in the
$45^\circ$ slope transition. The figure summarizes the posture-dependent load
moment, limited execution margin near $\SD(\gamma)$, and delay and uncertainty
sources in the hardware feedback loop.}
\label{fig:supp-hardware}
\end{figure}

\subsubsection{Hardware Execution Margin}

The hardware result on the $45^\circ$ slope should be interpreted in terms of
closed-loop execution margin, not incline angle alone. With a lower fixed
suction anchor, the plane-to-slope transition can create a long distal moment
arm, so the supporting segment may operate close to the boundary of
$\SD(\gamma)$. Moreover, $\SD(\gamma)$ is calibrated for quasi-static
load--deformation feasibility and does not include pneumatic delay, pressure
hysteresis, sealing uncertainty, adhesion-dependent boundary variation, or
motion-capture latency. Therefore, a planned state may satisfy the
quasi-static feasibility check but still be difficult to track on hardware
when the realized state drifts toward a low-margin region.
This distinction is summarized in Fig.~\ref{fig:supp-hardware}. The
planner-side model captures gravity-conditioned quasi-static feasibility,
whereas the hardware loop introduces additional delay and uncertainty between
the desired deformation, pressure command, suction boundary condition, and
measured robot state. The observed difficulty in the $45^\circ$ transition
therefore does not by itself indicate a mismatch in the construction of
$\SD(\gamma)$; it mainly reflects the gap between quasi-static feasibility and
robust closed-loop trackability on the current prototype. A natural extension
is to incorporate feedback-trackability margins or to learn a closed-loop
feasible family from hardware executions.

\end{document}